\documentclass[runningheads]{llncs}

\usepackage{eccv}

\usepackage{eccvabbrv}
\usepackage{graphicx}
\usepackage{booktabs}
\usepackage{multirow}
\usepackage{amsmath,amssymb}
\usepackage{mathtools}
\usepackage{bm}
\usepackage{enumitem}
\usepackage[accsupp]{axessibility}
\usepackage{array}
\usepackage[table]{xcolor}
\usepackage{makecell}
\usepackage{array}
\usepackage{wrapfig}
\usepackage{mathtools}
\usepackage{ragged2e} 
\usepackage{float}
\usepackage{algorithm}
\usepackage{algpseudocode}
\usepackage{setspace}
\usepackage[numbers]{natbib}
\usepackage{threeparttable}

\usepackage{hyperref}
\hypersetup{
    colorlinks=true,
    linkcolor=red,
    citecolor=eccvblue,
    urlcolor=magenta,
}

\usepackage{orcidlink}
\usepackage{wrapfig}
\usepackage{needspace}
\usepackage{caption}
\usepackage{placeins}
\usepackage{ragged2e} 

\title{Prune Once: Retraining-Free Task-Agnostic Pruning for Vision-Language Models}
\titlerunning{PORTA}
\author{
Minseok Kang\inst{1}$^*$\orcidlink{0009-0009-0801-905X} \and
Hyunwoo Kim\inst{1}$^*$\orcidlink{0009-0006-7071-7711} \and
Chanyoung Kim\inst{2}\orcidlink{0009-0008-7351-3399} \and
Minwoo Kim\inst{1}\orcidlink{0009-0007-0381-0067} \and
Jaekoo Lee\inst{3}$^\dagger$\orcidlink{0000-0002-5947-5487} \and
Dahuin Jung\inst{1}$^\dagger$\orcidlink{0000-0002-1344-1054}
}

\authorrunning{Kang et al.}

\institute{
Department of Artificial Intelligence, Chung-Ang University, Seoul, Korea \and
School of Computer Science and Engineering, Soongsil University, Seoul, Korea \and
Department of Computer Science, Kookmin University, Seoul, Korea \\
\email{dahuinjung@cau.ac.kr}
}

\begin{document}
\maketitle
\setcounter{footnote}{0}
\renewcommand{\thefootnote}{}
\footnotetext{\hspace{-0.7em}$^*$Equal Contribution , \hspace{-0.7em}\quad $^\dagger$Corresponding Authors}
\renewcommand{\thefootnote}{\arabic{footnote}}

\begin{abstract}


Vision-language models (VLMs) have achieved remarkable generalization across diverse multimodal tasks through large-scale pretraining, yet their rapidly increasing computational and memory requirements pose significant challenges for deployment in constrained environments. Existing pruning strategies often depend on task-specific criteria or LLM-oriented importance measures, making them unsuitable for task-agnostic pruning, where no task-specific samples are available at pruning time and the pruned model remains broadly applicable. We introduce a retraining-free VLM pruning framework called PORTA that derives a task- and modality-agnostic importance formulation based on activation variation, estimated from generic calibration data, which reliably captures feature-level representation utility across modalities. PORTA further incorporates an adaptive sparsity allocation mechanism that assigns layer-wise pruning ratios based on output feature variability, avoiding the limitations of uniform sparsity and reducing performance degradation at high compression levels. Extensive experiments across VLM architectures, such as CLIP, BLIP, and \textcolor{black}{Qwen2-VL}, demonstrate that PORTA achieves competitive downstream performance under high sparsity without requiring any retraining, supporting efficient VLM compression. Code is available at \url{https://github.com/cau-hai-lab/PORTA.git}.

\end{abstract}




\section{Introduction}
\label{sec:intro}

Vision-language models (VLMs)~\cite{radford2021learning,li2022blip,li2023blip2} have established a unified framework for joint visual and textual representation learning, enabling a broad spectrum of multimodal understanding tasks. Through large-scale pretraining on web-scale image–text pairs, these models effectively capture cross-modal correspondences and exhibit strong generalization across downstream tasks such as retrieval~\cite{cao2022image} and \textcolor{black}{VQA~\cite{Antol2015vqa}}. Pioneering frameworks such as CLIP~\cite{radford2021learning} and BLIP~\cite{li2022blip} exemplify this progress, leveraging web-scale data to acquire transferable multimodal features that enable zero-shot adaptation to new domains and tasks, thereby unifying multiple vision-language capabilities within a single model.

Despite their impressive performance, the ever-increasing scale of VLMs in terms of parameters and computation has rendered their deployment highly challenging in resource-constrained environments such as edge devices and real-time systems. To mitigate these limitations, a broad spectrum of model compression techniques—such as quantization~\cite{han2015deep,williams2023does}, knowledge distillation~\cite{wang2022efficientvlm}, and pruning—have been actively investigated. Among these, pruning~\cite{lee2018,wang2020,alizadeh2022prospect,dejorge2021progressive,Tanaka2020PruningNN,wang2023ntksap,shi2023upop,farina2024multiflow,sun2023simple} has emerged as a prominent method that reduces network complexity by removing parameters with low contribution to task performance, leading to reduced computational cost while maintaining accuracy.\par
\begin{wrapfigure}{r}{0.5\columnwidth}
  \vspace{-1.5\baselineskip}
  \centering
  \includegraphics[width=\linewidth]{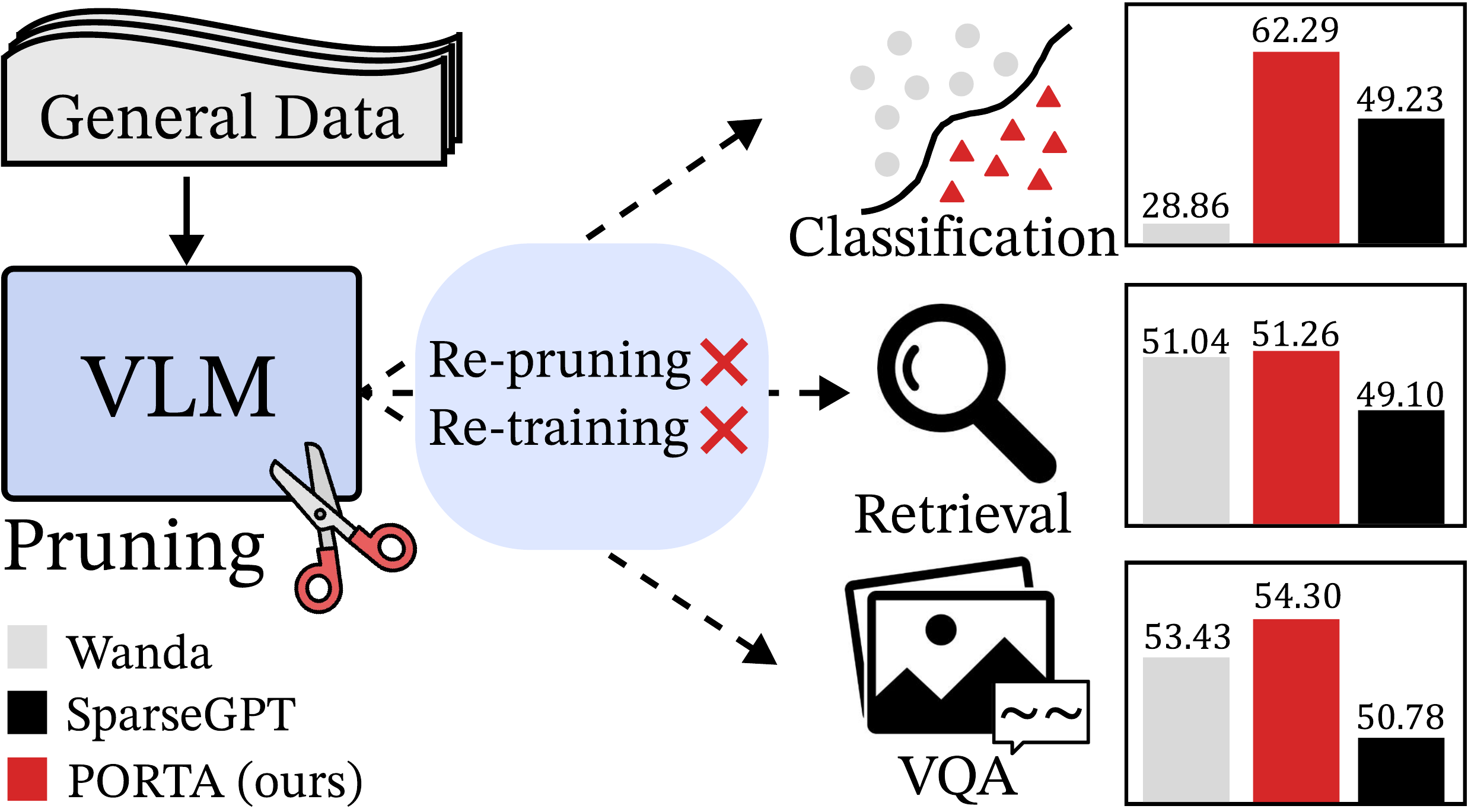}
  \caption{Illustration of PORTA's task- and modality-agnostic pruning paradigm: using only general data, a VLM is pruned once—without any re-pruning or retraining—and directly applied to diverse downstream tasks. \textcolor{black}{Comparison across three tasks—classification (accuracy), retrieval (TR@5), and VQA (Avg)—under high sparsity.} PORTA delivers uniformly strong results across all settings, outperforming Wanda and SparseGPT.}
  \label{fig:fig1}
  \vspace{-1.7\baselineskip}
\end{wrapfigure}
\begin{figure*}[t]
  \centering
  \includegraphics[width=\textwidth]{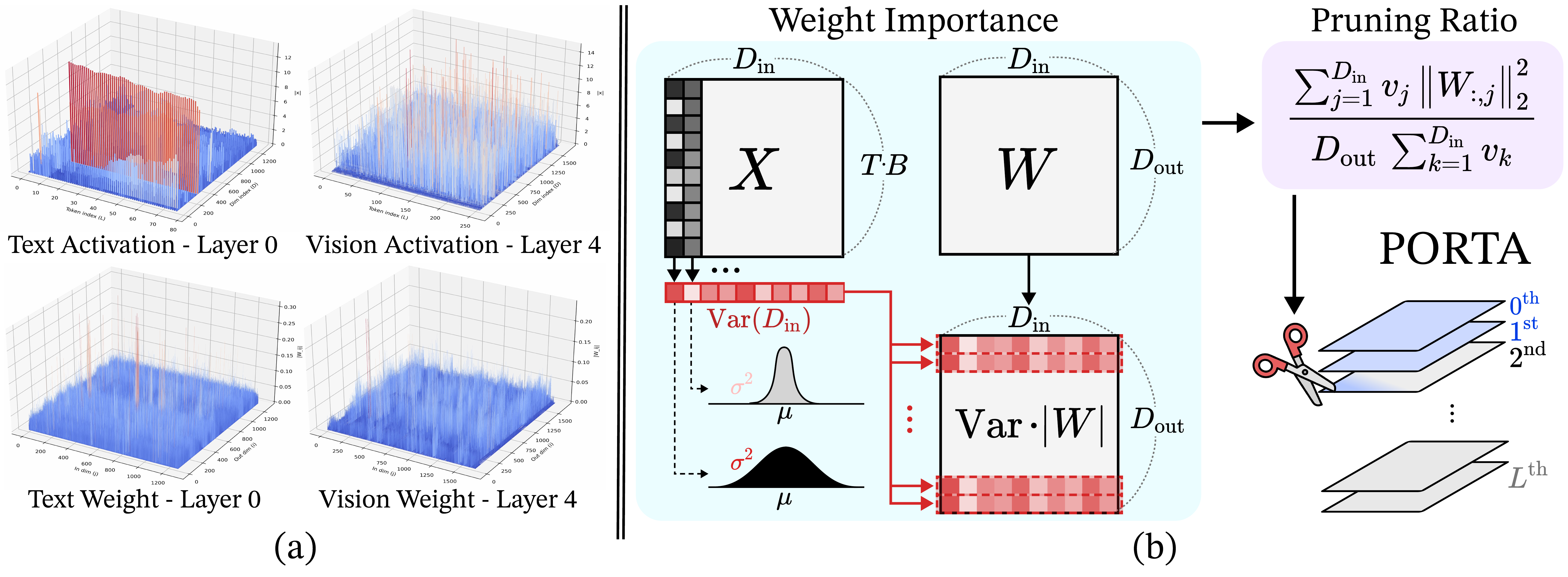}
  \caption{Modality-specific activation characteristics and PORTA overview. The left component, (a) Text and vision branches exhibit clearly different activation patterns across layers. 
(b) PORTA overview: $\mathrm{Var}\!\cdot\!|W|$ for weight importance and a pruning-ratio score for layer-wise sparsity allocation.}
  \label{fig:fig2}
  \vspace{-1.5\baselineskip}
\end{figure*}

\textcolor{black}{Task-agnostic pruning for VLMs —where a model is pruned once and reused across diverse downstream tasks without re-pruning or retraining, as illustrated in Fig.~\ref{fig:fig1}—remains unexplored despite its practical importance. Most existing pruning methods} rely on task-specific criteria, requiring re-pruning or re-training whenever the downstream task changes, which introduces non-negligible overhead~\cite{blalock2020state}.
\textcolor{black}{In particular, in VLM pruning, many approaches adopt LLM-based weight-importance measures such as Wanda~\cite{sun2023simple} and SparseGPT~\cite{frantar2023sparsegpt}.} \textcolor{black} {However, recent work shows that the resulting importance estimates can be sensitive to the choice of calibration data~\cite{williams2023does}. Consequently, in task-agnostic settings—where task-specific calibration samples are difficult to obtain—importance estimation may become unstable, limiting the applicability of such methods.
Moreover, Wanda is motivated by activation outlier patterns observed in LLMs~\cite{sun2023simple}. As illustrated in Fig.~\ref{fig:fig2} (a), these patterns do not manifest in the same manner in vision models, which may reduce their effectiveness when applied to VLMs. This observation suggests that pruning criteria derived from activation characteristics in LLMs may not generalize uniformly across the vision and language components of VLMs. In particular, when importance estimation relies heavily on modality-specific activation distribution properties, it may introduce modality bias, causing pruning to be disproportionately concentrated in either the vision or language branch. This issue becomes more pronounced in task-agnostic scenarios, where the absence of task-specific calibration signals prevents effective mitigation of such bias.}



To address this bias, we propose PORTA (Prune-Once: Retraining-free Task-Agnostic pruning), which does not simply replace existing weight-importance statistics but introduces a new importance formulation that can reflect the characteristics of VLMs. \textcolor{black}{PORTA utilizes activation variation, which remains comparatively stable across modalities, to capture each feature’s representation utility and construct a task- and modality-agnostic importance formulation. Specifically, PORTA derives feature importance from the variation of activations along the feature dimension, interpreting broader fluctuations as evidence of stronger engagement across diverse inputs and higher representation utility.} This formulation mitigates modality-induced bias, maintains coherent feature-space behavior by grounding pruning decisions in the model’s representational behavior, and enables seamless transfer across heterogeneous tasks \textcolor{black}{without re-pruning, as its importance estimates are robust to the choice of calibration data}, achieving improved efficiency across modalities.
\textcolor{black}{Concretely, PORTA incorporates two components to enable layer-wise sparsity allocation: a scoring criterion for selecting important features and a ratio mechanism that adaptively allocates layer-wise sparsity across network blocks.} The scoring criterion estimates feature importance from the activation variation along the feature axis—low-variability features that react only to specific contexts are pruned, while high-variability features with broader responsiveness are retained. The ratio mechanism assigns different sparsity levels to each layer according to the variability of its output feature representation, measured from \textcolor{black}{activation variation,} thereby alleviating the limitation of layer-uniform pruning. Layers exhibiting higher \textcolor{black}{representation utilization} are assigned lower sparsity, thereby mitigating performance cliffs through selective pruning of less contributive features.

As shown in Fig.~\ref{fig:fig1}, across extensive zero-shot evaluations, PORTA consistently delivers strong pruning performance on a wide range of VLM architectures and tasks. On CLIP, PORTA achieves substantial gains over Wanda, SparseGPT, ECoFLaP~\cite{sung2024ecoflap}, and Multiflow~\cite{farina2024multiflow} in image classification and image–text retrieval, particularly at high sparsity levels where alternative methods exhibit severe degradation. \textcolor{black}{Moreover, PORTA generalizes beyond CLIP-based contrastive VLMs to generative multimodal architectures such as \textcolor{black}{Qwen2-VL~\cite{wang2024qwen2vl}}, while preserving strong downstream performance on \textcolor{black}{VQA~\cite{Antol2015vqa}} even after pruning.} \textcolor{black}{Our experiments confirm that PORTA maintains strong downstream performance across architectures, tasks, and sparsity levels, demonstrating that task- and modality-agnostic pruning can be achieved without re-pruning, as the proposed importance estimation remains robust to the choice of calibration data.} The main contributions of this work are summarized as follows:
\begin{itemize}
    \item \textcolor{black}{We propose PORTA, a prune-once, retraining-free pruning framework for VLMs. PORTA prunes a pretrained model once and transfers it across diverse downstream tasks without any additional training.}
    \item We introduce a modality-agnostic importance criterion and an adaptive sparsity allocation mechanism. The criterion quantifies feature salience via activation-based statistical variability, reflecting each feature’s contribution to representational capacity, while sparsity ratios are adaptively assigned based on the variability of the output feature space at each layer.
    \item \textcolor{black}{We demonstrate that PORTA achieves robust performance across multiple VLM architectures and sparsity levels. In particular, PORTA maintains competitive accuracy compared with existing pruning methods even under high sparsity, validating its robustness and efficiency in large-scale multimodal compression.}
\end{itemize}

\section{Related Works}
\label{sec:relatedworks}
\subsection{Vision-Language Models}
Vision-language models (VLMs) that jointly learn visual and textual representations have been actively explored in recent years~\cite{radford2021learning,li2022blip,li2023blip2,liu2023visual}. Among them, CLIP~\cite{radford2021learning} demonstrates that a single model can be broadly applied to various vision-language tasks such as classification, retrieval, and \textcolor{black}{VQA}, by aligning visual and textual representations in a shared embedding space through contrastive learning on large-scale image–text pairs. Building on this paradigm, BLIP~\cite{li2022blip} connects a pretrained vision encoder with a language model to enable more general-purpose vision-language generation and understanding. Recent studies further expand model capacity to capture more complex multimodal information, yet the resulting increase in parameters and computational demand poses challenges for deployment in resource-constrained environments.
\subsection{Post-Training Pruning}
As models continue to scale, the computational cost of iterative pruning has become prohibitive, motivating the adoption of one-shot pruning as a more practical alternative. Representative approaches include SparseGPT~\cite{frantar2023sparsegpt} and Wanda~\cite{sun2023simple}, which leverages the emergent large-magnitude activations observed in LLMs.
However, when applied to VLMs, these strategies reveal notable limitations. \textcolor{black}{Recent findings \cite{williams2023does} indicate that both SparseGPT and Wanda are highly sensitive to the selection of calibration data. In task-agnostic VLM scenarios—where task-specific calibration data cannot be utilized—this sensitivity results in suboptimal pruning outcomes. Furthermore, Wanda’s reliance on activation norms makes it sensitive to the differing activation magnitudes of VLMs, making it less appropriate for multimodal pruning.}

Recent studies have attempted to design pruning strategies specifically for VLMs, such as Multiflow~\cite{farina2024multiflow} and ECoFLaP \cite{sung2024ecoflap}. Multiflow focuses on information flow by leveraging modality-specific weight distributions for task-agnostic pruning, whereas ECoFLaP introduces an efficient layer-wise ratio estimation method using zeroth-order gradients to reduce computational overhead. Nonetheless, Multiflow still requires task-specific fine-tuning, incurring additional cost, and ECoFLaP does not introduce its own importance metric; instead, it adopts Wanda, whose assumptions are rooted in LLM behavior and are not directly aligned with VLMs.

\subsection{Task-Agnostic Pruning}

In continual learning~\cite{he2020task} and few-shot learning~\cite{jamal2019task}, task-agnostic approaches aim to acquire representations or update rules that generalize to unseen tasks, emphasizing robustness to task shifts rather than reliance on task-specific supervision. Beyond learning, the notion of “task-agnostic” has gained renewed importance in the context of model compression for large models~\cite{zhang2022bmcook,xu2021bert,wang2023task}. Among them, Multiflow~\cite{farina2024multiflow} highlights that VLMs are typically deployed across heterogeneous downstream tasks, and that using task-specific calibration data for pruning undermines generalization. Accordingly, Multiflow formalizes task-agnostic pruning as performing pruning without access to downstream labels or task-specific samples, ensuring that the pruned model remains broadly usable across retrieval, classification, and generation tasks. In this landscape, PORTA advances task-agnostic pruning by performing pruning without any form of retraining, achieving stable performance across modalities and tasks.

\section{PORTA}
In this section, we present the motivation and key components of PORTA, our pruning framework designed for vision–language models (VLMs). PORTA aims to provide a pruning mechanism that remains robust to variations in calibration data and modality-specific activation characteristics—two limitations that hinder the reliability of existing VLM pruning approaches. To address these limitations, PORTA introduces two components. First, we develop a novel importance formulation that mitigates input-distribution sensitivity in multimodal pruning settings by leveraging feature activation variance, enabling a reliable assessment of each feature’s representation utility. Second, we propose a layer-wise pruning ratio estimation strategy that accounts for architectural heterogeneity in VLMs by evaluating the variability of layer output feature spaces. An overview of PORTA is provided in Fig.~\ref{fig:fig2} (b).

\subsection{Variance as a Modality-Agnostic Importance Score}\label{sec:3.1}

\begin{table}[t]
\begin{minipage}[t]{0.48\linewidth}
\centering
\scriptsize
\setlength{\tabcolsep}{6pt}
\renewcommand{\arraystretch}{1.2}
\caption{Layer-wise max–min dispersion of activation statistics under magnitude- and variance-based measures. Variance exhibits reduced modality sensitivity.}
\label{tab:modality_var}
\begin{tabular}{c c c c c}
\toprule
\textbf{Method} & \textbf{Module} & \textbf{L1} & \textbf{L20} & \textbf{L30} \\
\midrule
Mag. & Text   & 3.44 & 7.11 & 7.07 \\
Mag. & Vision & 4.38 & 4.85 & 3.73 \\
Var. & Text   & 3.67 & 4.48 & 4.61 \\
Var. & Vision & 3.73 & 4.47 & 4.11 \\
\bottomrule
\end{tabular}
\end{minipage}
\hfill
\begin{minipage}[t]{0.48\linewidth}
\centering
\scriptsize
\setlength{\tabcolsep}{5pt}
\renewcommand{\arraystretch}{1.3}
\caption{Zero-shot retrieval accuracy after removing a percentage of features based on variance ranking.}
\label{tab:toy}
\begin{tabular}{lccc}
\toprule
\textbf{Removed Features} & \textbf{5\%} & \textbf{10\%} & \textbf{15\%} \\
\midrule
High-Variance   & 0.02  & 0.02  & 0.02  \\
Random-Variance & 67.00 & 63.62 & 47.60 \\
Low-Variance    & 68.34 & 67.88 & 67.28 \\
\bottomrule
\end{tabular}
\end{minipage}
\vspace{-0.5em}
\end{table}

Existing pruning methods predominantly rely on activation magnitude as the primary criterion for importance estimation. 
This assumption is largely motivated by activation outlier patterns observed in large language models, where dimensions exhibiting extreme responses are often considered highly influential. However, such magnitude-based assumptions may not directly transfer to multimodal architectures such as VLMs. 
In VLMs, the activation scale and distribution characteristics differ substantially between the vision and language encoders. 
In particular, pronounced outlier patterns commonly observed in text representations are less evident in vision features. Such modality-dependent differences can also be observed in Fig.~\ref{fig:fig2} (a).
As a result, magnitude-based importance estimation may introduce modality-dependent bias, potentially leading to imbalanced pruning across modalities.
To analyze modality sensitivity, we examine magnitude- and variance-based activation statistics, and then analyze the relationship between feature variance and representation utility.

\subsubsection{Modality Sensitivity of Activation Statistics.}

We analyze the modality sensitivity of activation statistics by measuring their dispersion across layers and modalities.
For each layer, we compute two quantities: 
(i) the range of dimension-wise mean activations, defined as the difference between the maximum and minimum mean activation values across feature dimensions; and 
(ii) the range of dimension-wise activation variances, defined as the difference between the maximum and minimum variance values across feature dimensions.

Tab.~\ref{tab:modality_var} reports these max–min ranges under magnitude- and variance-based statistics.
As shown, magnitude-based statistics exhibit substantial variation across layers and between modalities, reflecting strong sensitivity to modality-specific activation scales.
In contrast, variance-based statistics demonstrate comparatively stable dispersion across both vision and language encoders, indicating reduced modality dependence.
These findings suggest that variance provides a more stable and modality-agnostic signal than magnitude in multimodal settings, motivating its use as the foundation of our importance formulation.

\subsubsection{Representation Utility and Feature
Variance.}
We consider feature variance as an alternative indicator of representation utility. Intuitively, features with larger variance tend to respond broadly across tokens or patches and thus capture generalizable structures, whereas low-variance features often encode localized or input-specific patterns.

To empirically validate the relationship between feature variance and representation utility, we conduct an axis-removal analysis based on feature variance in Tab.~\ref{tab:toy}. 
Specifically, we compare three removal schemes: 
(i) High-Variance features, 
(ii) Random features, and 
(iii) Low-Variance features. 

For each scheme, we remove a certain percentage of feature dimensions according to the variance ranking while keeping all other model parameters fixed, and evaluate zero-shot retrieval performance.

Across all tested removal ratios, removing high-variance features causes catastrophic performance degradation, whereas removing low-variance features results in minimal accuracy drop. 
These results indicate that high-variance features carry most of the representational utility, validating variance as a meaningful importance signal.

The results indicate that low-variance features can be removed with minimal impact on performance and variance provides a robust and modality-agnostic basis for importance estimation in multimodal pruning.
Unlike magnitude-based criteria that focus on extreme responses, variance reflects the breadth of representation utilization across inputs, providing a balanced measure of contribution.

Based on the above analysis, feature variance is incorporated into the weight tensor following the standard weight-pruning paradigm~\cite{han2015learning,sun2023simple}.

\subsection{Weight Importance}\label{sec:3.2}
We quantify representation utility for each layer by measuring the variance of its input
features along the token/patch dimension.  
Let the input activations to layer $l$ be 
$X \in \mathbb{R}^{B \times T \times D_{\mathrm{in}}}$ and the corresponding weight matrix be 
$W \in \mathbb{R}^{D_{\mathrm{out}} \times D_{\mathrm{in}}}$.  
For each input feature $j$, we compute its variability as
\begin{equation}
    v_j = \mathrm{Var}_{b,t}\!\left(X_{b,t,j}\right), 
    \qquad j = 1,\dots, D_{\mathrm{in}},
\end{equation} 
where $v_j$ serves as a measure of representation utility:  
features exhibiting low variance respond only to a narrow subset of input tokens or patches and thus encode highly specific, localized patterns, whereas high-variance features participate broadly across inputs and therefore capture generalizable structure.

To incorporate representation utility into pruning, we combine feature variance with weight magnitude.
The variance vector $v$ is broadcast along the output dimension and multiplied with the absolute
weight matrix, yielding the final importance score
\begin{equation}
    S_{i,j} = v_j \cdot \lvert W_{i,j} \rvert,
    \qquad i = 1,\dots,D_{\mathrm{out}}, \ j = 1,\dots,D_{\mathrm{in}},
\end{equation}
where $S_{i,j}$ reflects both the scale of each parameter and the extent to which the corresponding input feature contributes to the model’s representational capacity.

\subsection{Pruning Ratio} \label{sec:3.3}
Different layers within deep models are known to play distinct roles and contribute unequally to the resulting representations~\cite{yin2023outlier,wang2024exploring,lee2021layeradaptive}. In VLMs, this variation is further influenced by their structurally distinct architectural components and modality-specific fusion structures, which lead layers to participate differently in forming multimodal features. Since such architectures do not impose a fixed modality weighting or uniform layer structure, assigning a consistent pruning ratio across layers requires a measure that is both modality- and task-agnostic. To achieve a balanced allocation of pruning across diverse structures, PORTA measures each layer output's feature-space utilization using its output variance, which serves as an indicator of how much representational capacity the layer occupies. 

To define and introduce the pruning-ratio score for each layer, we first establish the notation. Let $X \in \mathbb{R}^{T \times D_{\text{in}}}$ denote the input tokens, 
where each row $x_t$ represents a token, and let 
$W \in \mathbb{R}^{D_{\text{out}} \times D_{\text{in}}}$ be the weight matrix of a linear layer. 
The layer output is given by
\begin{equation}
    Y = X W^\top \in \mathbb{R}^{T  \times D_{\text{out}}}.
\end{equation}
Denoting the $o$-th row of $W$ by $w_o \in \mathbb{R}^{1 \times D_{\text{in}}}$, 
the $o$-th output channel (the $o$-th column of $Y$) can be expressed as
\begin{equation}
    y_{:,o} = X w_o^\top.
\end{equation}
For analytical simplicity, we consider each input token $x_t$ is drawn from a zero-mean distribution with input covariance $\Sigma_X = \mathbb{E}[x_t^\top x_t]$, the variance of the $o$-th output channel across the token dimension then becomes
\begin{align}
    \operatorname{Var}[y_{:,o}]
    &\approx \mathbb{E}_{x_t}\bigl[(x_t w_o^\top)^2\bigr] \\
    &= \mathbb{E}_{x_t}\bigl[x_t w_o^\top w_o x_t^\top\bigr] \\
    &= \mathbb{E}_{x_t}\bigl[\operatorname{Tr}(w_o^\top w_o x_t^\top x_t)\bigr] \\
    &= \operatorname{Tr}\bigl(w_o^\top w_o\, \Sigma_X\bigr)
     \;=\; w_o \Sigma_X w_o^\top.
\end{align}
We propose a layer-wise score for determining pruning ratios in PORTA.  
To eliminate scale effects and ensure fair comparison across layers,
we first normalize the input covariance so that only its directional
structure is retained:
\begin{equation}
    \widehat{\Sigma}_X \;\coloneqq\;
    \frac{\Sigma_X}{\operatorname{Tr}(\Sigma_X)} .
\end{equation}
Based on normalized covariance, the pruning ratio for each layer is
determined using the following score:
\begin{equation}
    s_{\text{layer}}
    \;\coloneqq\;
    \frac{1}{D_{\text{out}}}
    \sum_{o=1}^{D_{\text{out}}}
        w_o \,\widehat{\Sigma}_X\, w_o^\top ,
\label{eq:ratio-score}
\end{equation}
where averaging over output channels ensures consistency across layers
with different output dimensionalities.

By stacking all output weight vectors $w_o$ into a matrix $W=[w_1;\dots;w_{D_{\text{out}}}]$, we equivalently write Eq.~\ref{eq:ratio-score} as follows:
\begin{equation}
    s_{\text{layer}}
    =
    \frac{1}{D_{\text{out}}}\,
    \operatorname{Tr}\!\bigl(W^\top W\,\widehat{\Sigma}_X\bigr)
    =
    \frac{1}{D_{\text{out}}}
    \sum_{j=1}^{D_{\text{in}}}
    \bigl\|\bigl(W\,\widehat{\Sigma}_X^{1/2}\bigr)_{:,j}\bigr\|_2^2.
\end{equation}

\begin{algorithm}[t]
  \caption{Layer-wise Pruning Ratio Allocation}
  \label{alg:s_layer_ratio}
  \small
  \setstretch{1.08} 
  \begin{algorithmic}[1]
    \Require $\{s_i\}_{i\in\mathcal L},\ S\in(0,1),\ \alpha>0,\ \{w_i\}_{i\in\mathcal L}$
    \Ensure  $\{r_i\}_{i\in\mathcal L}$

    \State $T \gets \sum_{i\in\mathcal L} s_i$ \Comment{sum over layer importance scores}

    \State $p_i \gets s_i/T$, for all $i\in\mathcal L$ \Comment{importance proportion}
    \State $u_i \gets (1-p_i)^\alpha$, for all $i\in\mathcal L$ \Comment{pruning score with strength $\alpha$}

    \State $W \gets \sum_i w_i$ 
    \State $\bar{u} \gets (\sum_i w_i u_i)/W$ \Comment{weight-adjusted mean pruning score}
    \State $C \gets S/\bar{u}$
    \State $r_i \gets C\,u_i$, for all $i\in\mathcal L$ \Comment{per-layer pruning ratio}

    \State \Return $\{r_i\}_{i\in\mathcal L}$
  \end{algorithmic}
\end{algorithm}
\vspace{-1.0\baselineskip}

As illustrated in Fig.~\ref{fig:fig2} (b), the formulation can be interpreted as the average squared $\ell_2$ norm of the input-direction feature vectors of $W$ projected onto the input feature covariance space. 
The layer output feature variance score reflects the joint contribution of the input feature normalized covariance $\widehat{\Sigma}_X$and the $\ell_2$  magnitude of the weight vectors along each input direction.

We approximate the covariance $\Sigma_X$ with its diagonal for computational and memory
efficiency, $\Sigma_X \approx \mathrm{diag}(v_1,\dots,v_{D_{\text{in}}})$ with $v_j = \operatorname{Var}[x_{t,j}]$. In variance-only case,
$s_{\text{layer}}$ reduces to
\begin{equation}
    s_{\text{layer}}
    \;=\;
    \frac{   \sum_{j=1}^{D_{\text{in}}} v_j \,\bigl\| W_{:,j} \bigr\|_2^2}{D_{\text{out}}\,\sum_{k=1}^{D_{\text{in}}} v_k},
    \qquad v_j=\operatorname{Var}[x_{t,j}],
\end{equation}
where each column vector $W_{:,j}$ quantifies the influence of the $j$-th input feature on the output channels, while the variance term $v_j$ measures the representation utility of the input feature. The resulting score $s_{\text{layer}}$ represents a normalized, variance-weighted
average of $\|W_{:,j}\|_2^2$, assigning larger contribution to input dimensions that both exhibit higher utility and exert stronger effect on the layer’s output. Layers whose output representations rely heavily on such high-utility directions receive larger scores and are therefore allocated lower pruning ratios. The computed $s_{\text{layer}}$ values are further normalized across layers to satisfy a target global sparsity, as detailed in Algorithm~\ref{alg:s_layer_ratio}.

\section{Experiments}
In this section, we conduct zero-shot evaluations of PORTA across three task categories—image classification, image–text retrieval, and  \textcolor{black}{VQA}—using three representative VLMs: \textcolor{black}{OpenAI CLIP-ViT-bigG, BLIP-base~\cite{li2022blip}, and \textcolor{black}{Qwen2-VL~\cite{wang2024qwen2vl}.}} All evaluations are performed in a fully zero-shot setting using publicly released pretrained checkpoints.
\subsection{Experimental Setup}
\label{sec:4.1}
We evaluate PORTA under zero-shot settings across three task families: image classification, image–text retrieval, and \textcolor{black}{VQA.} For classification, we use four benchmarks—CIFAR-10~\cite{krizhevsky2009cifar}, CIFAR-100~\cite{krizhevsky2009cifar}, ImageNet-1K~\cite{deng2009imagenet}, and Oxford Flower-102~\cite{nilsback2008automated}—to assess performance across datasets with varying granularity and scale. Image–text retrieval experiments are conducted on MSCOCO~\cite{lin2014microsoft} and Flickr30K~\cite{plummer2015flickr30k}. For MSCOCO, we follow the Karpathy split and report Recall@K for both image-to-text and text-to-image retrieval on the Karpathy test split, while Flickr30K results are obtained using its official test split. \textcolor{black}{For VQA, we use ScienceQA~\cite{lu2022learn} and report test-split accuracy across its subject, context-modality, and grade categories, along with the overall average.}\par
\definecolor{PORTAcolor}{HTML}{D9D9D9}
\begin{table*}[t]
\centering
\scriptsize
\caption{Zero-shot evaluation on image--text retrieval and image classification using CLIP pruned at 55\%, 60\%, and 65\% sparsity. PORTA consistently achieves strong performance across both retrieval (MSCOCO) and classification benchmarks, particularly under high sparsity, where alternative pruning methods exhibit substantial degradation. \textcolor{black}{C10: CIFAR-10, C100: CIFAR-100, IN: ImageNet, F102: Flowers102}.}
\renewcommand{\arraystretch}{1.2}
\setlength{\tabcolsep}{5pt}
\resizebox{1.0\linewidth}{!}{%
\begin{tabular}{
  >{\centering\arraybackslash}m{1.1cm}
  >{\centering\arraybackslash}m{1.7cm}
  @{\hspace{2pt}}
  *{4}{>{\centering\arraybackslash}m{0.75cm}}
  *{4}{>{\centering\arraybackslash}m{0.85cm}}
}
\toprule
\multirow{4}{*}{\textbf{Sparsity}} & \multirow{4}{*}{\textbf{Method}}
& \multicolumn{4}{c}{\textbf{Retrieval}}
& \multicolumn{4}{c}{\textbf{Classification}} \\
\cmidrule(lr){3-6}\cmidrule(lr){7-10}
& & \multicolumn{4}{c}{\textbf{MSCOCO}}
& \textbf{C10} & \textbf{C100} & \textbf{IN} & \textbf{F102} \\
\cmidrule(lr){3-6}\cmidrule(lr){7-10}
& & \textbf{TR@1} & \textbf{TR@5} & \textbf{IR@1} & \textbf{IR@5}
& \multicolumn{4}{c}{\textbf{Accuracy}} \\
\midrule
0\% & \textbf{Dense}
& 68.56 & 87.70 & 52.28 & 76.09 & 97.04 & 87.50 & 78.45 & 80.19 \\
\midrule
\multirow{5}{*}{55\%} & Wanda
& \textbf{66.30} & 86.44 & 48.62 & 73.05 & \textbf{96.12} & 79.92 & 69.50 & 67.72 \\
& SparseGPT
& 65.78 & 86.44 & 48.57 & \textbf{73.29} & 95.07 & 79.95 & 68.61 & 65.16 \\
& ECoFLaP
& 51.14 & 76.04 & 33.14 & 58.44 & 80.69 & 51.79 & 43.11 & 30.00 \\
& Multiflow
& 0.02 & 0.12 & 0.02 & 0.12 & 9.25 & 1.23 & 0.16 & 1.50 \\
\rowcolor{PORTAcolor}& \textbf{PORTA}
& 66.28 & \textbf{87.00} & \textbf{48.70} & 73.02 & 95.58 & \textbf{80.27} & \textbf{70.29} & \textbf{68.73} \\
\midrule
\multirow{5}{*}{60\%} & Wanda
& 59.78 & 82.82 & 42.55 & 68.01 & 93.20 & 70.28 & 57.35 & 51.48 \\
& SparseGPT
& 57.52 & 81.88 & 41.77 & 66.97 & \textbf{95.16} & 74.23 & 55.26 & 41.37 \\
& ECoFLaP
& 35.44 & 60.56 & 19.12 & 39.46 & 57.98 & 32.60 & 25.73 & 14.21 \\
& Multiflow
& 0.04 & 0.16 & 0.01 & 0.08 & 9.79 & 1.10 & 0.08 & 0.49 \\
\rowcolor{PORTAcolor}& \textbf{PORTA}
& \textbf{60.14} & \textbf{83.04} & \textbf{43.11} & \textbf{68.18} & 95.06 & \textbf{75.59} & \textbf{59.66} & \textbf{55.50} \\
\midrule
\multirow{5}{*}{65\%} & Wanda
& 27.30 & 51.04 & 16.96 & 35.61 & 69.81 & 28.86 & 30.06 & 17.30 \\
& SparseGPT
& 25.10 & 49.10 & 15.93 & 35.00 & 86.35 & 49.23 & 25.96 & 12.09 \\
& ECoFLaP
& 13.60 & 29.62 & 7.22 & 18.32 & 33.18 & 16.70 & 10.20 & 6.66 \\
& Multiflow
& 0.02 & 0.12 & 0.02 & 0.10 & 8.90 & 0.96 & 0.12 & 0.81 \\
\rowcolor{PORTAcolor}& \textbf{PORTA}
& \textbf{28.14} & \textbf{51.26} & \textbf{17.31} & \textbf{37.32} & \textbf{90.49} & \textbf{62.29} & \textbf{31.38} & \textbf{18.27} \\
\bottomrule
\end{tabular}
}
\label{tab:table4a_accuracy_8metrics}
\vspace{-1.0\baselineskip}
\end{table*}

\textcolor{black}{We assess PORTA on three representative VLMs using publicly available pretrained checkpoints without any fine-tuning: OpenAI CLIP-ViT-bigG and BLIP-base for classification and retrieval, and \textcolor{black}{Qwen2-VL~\cite{wang2024qwen2vl} for VQA.} For comparison, we include four pruning baselines: Wanda~\cite{sun2023simple}, which combines activation statistics with weight magnitudes; SparseGPT~\cite{frantar2023sparsegpt}, which uses Hessian-based approximations for one-shot pruning; ECoFLaP~\cite{sung2024ecoflap}, a zeroth-order pruning method tailored for VLMs; and Multiflow~\cite{farina2024multiflow}, a task-agnostic pruning approach built upon modality-specific weight distributions. Following our zero-shot protocol, we report Multiflow results without any task-specific fine-tuning; additional comparisons with Multiflow fine-tuning are provided in Appendix \textcolor{red}{A}.}
\textcolor{black}{For calibration, we estimate activation variation from a general image--text set. We use MSCOCO as a standard, publicly available calibration source, which contains 82{,}783 image--text pairs. Importantly, PORTA does not rely on large-scale or downstream-aligned calibration.}

All experiments are conducted on a single NVIDIA L40S GPU. We focus on the standard one-shot pruning setting and evaluate sparsity levels of $45\%$, $50\%$, $55\%$, $60\%$, and $65\%$. To ensure a fair comparison across pruning methods, all evaluations use identical calibration data and identical sample counts, so that performance differences arise solely from the pruning strategies.













\definecolor{PORTAcolor}{HTML}{D9D9D9}
\begin{table*}[t]
\centering
\scriptsize
\caption{Zero-shot image--text retrieval results on Flickr30k using BLIP at 45\% sparsity. PORTA attains the highest mean performance among pruning baselines.}
\renewcommand{\arraystretch}{1.2}
\setlength{\tabcolsep}{5pt}
\resizebox{1.0\textwidth}{!}{%
\begin{tabular}{
  >{\centering\arraybackslash}m{1.1cm}
  >{\centering\arraybackslash}m{1.9cm}
  >{\centering\arraybackslash}m{0.8cm}
  >{\centering\arraybackslash}m{0.8cm}
  >{\centering\arraybackslash}m{0.8cm}
  >{\centering\arraybackslash}m{1.2cm}
  >{\centering\arraybackslash}m{0.8cm}
  >{\centering\arraybackslash}m{0.8cm}
  >{\centering\arraybackslash}m{0.8cm}
  >{\centering\arraybackslash}m{1.2cm}
  >{\centering\arraybackslash}m{0.9cm}
}
\toprule
\multirow{4}{*}{\textbf{Sparsity}} & \multirow{4}{*}{\textbf{Method}}
& \multicolumn{9}{c}{\textbf{Image-Text Retrieval}} \\
\cmidrule(lr){3-11}
& & \multicolumn{9}{c}{\textbf{Flickr30k}} \\
\cmidrule(lr){3-11}
& & \textbf{TR@1} & \textbf{TR@5} & \textbf{TR@10} & \textbf{TR@mean}
  & \textbf{IR@1} & \textbf{IR@5} & \textbf{IR@10} & \textbf{IR@mean} & \textbf{Mean} \\
\midrule
0\% & \textbf{Dense}
& 83.10 & 96.60 & 97.90 & 92.53 & 78.48 & 93.82 & 96.52 & 89.61 & 91.07 \\
\midrule
\multirow{5}{*}{45\%} & Wanda
& 82.20 & 95.70 & 98.10 & 92.00 & 74.40 & 92.82 & 95.66 & 87.63 & 89.81 \\
& SparseGPT
& 81.70 & 95.10 & \textbf{98.20} & 91.67 & \textbf{75.86} & 93.02 & \textbf{95.82} & \textbf{88.23} & 89.95 \\
& ECoFLaP
& 79.20 & 93.20 & 96.40 & 89.60 & 69.78 & 90.50 & 94.66 & 84.98 & 87.29 \\
& Multiflow
& 79.50 & 94.20 & 97.00 & 90.43 & 74.48 & 92.50 & 95.48 & 87.49 & 88.96 \\
\rowcolor{PORTAcolor} & \textbf{PORTA}
& \textbf{83.00} & \textbf{96.00} & 98.00 & \textbf{92.33} & 75.62 & \textbf{93.10} & \textbf{95.82} & 88.18 & \textbf{90.26} \\
\bottomrule
\end{tabular}
}
\label{tab:table3_retrieval_flickr30k}
\vspace{0\baselineskip}
\end{table*}
\begin{table*}[t]
\centering
\scriptsize
\caption{Zero-shot VQA results on ScienceQA at 50\% sparsity with Qwen2-VL. PORTA achieves the highest average accuracy among pruning baselines.}
\renewcommand{\arraystretch}{1.2}
\setlength{\tabcolsep}{6pt}
\resizebox{\textwidth}{!}{%
\begin{tabular}{
  >{\centering\arraybackslash}m{0.9cm}     
  >{\centering\arraybackslash}m{2.4cm}     
  @{\hspace{2pt}}
  >{\centering\arraybackslash}m{0.8cm}     
  >{\centering\arraybackslash}m{0.8cm}     
  >{\centering\arraybackslash}m{0.8cm}     
  >{\centering\arraybackslash}m{0.8cm}     
  >{\centering\arraybackslash}m{0.8cm}     
  >{\centering\arraybackslash}m{0.8cm}     
  >{\centering\arraybackslash}m{0.8cm}     
  >{\centering\arraybackslash}m{1.0cm}     
  >{\centering\arraybackslash}m{1.0cm}     
}
\toprule
\multirow{3}{*}[2pt]{\textbf{Sparsity}} & \multirow{3}{*}[2pt]{\textbf{Method}}
& \multicolumn{3}{c}{\textbf{Subject}}
& \multicolumn{3}{c}{\textbf{Context Modality}}
& \multicolumn{2}{c}{\textbf{Grade}}
& \multirow{3}{*}[2pt]{\textbf{Average}} \\
\cmidrule(lr){3-5}\cmidrule(lr){6-8}\cmidrule(lr){9-10}
& & \textbf{NAT} & \textbf{SOC} & \textbf{LAN}
  & \textbf{TXT} & \textbf{IMG} & \textbf{NO}
  & \textbf{G1-6} & \textbf{G7-12} & \\
\midrule
0\% & \textbf{Dense}
& 60.66 & 67.27 & 60.00 & 62.37 & 60.98 & 73.77 & 65.57 & 55.24 & 61.87 \\
\midrule
\multirow{5}{*}{50\%}
& Wanda
& 55.01 & 47.35 & 55.09 & 56.03 & 49.97 & \textbf{75.40} & 55.80 & 49.17 & 53.43 \\
& SparseGPT
& 49.42 & 49.04 & 55.00 & 53.16 & 47.74 & 67.21 & 53.56 & 45.74 & 50.78 \\
& ECoFLaP
& 51.33 & \textbf{50.95} & 52.19 & 52.61 & 49.88 & 63.93 & 54.52 & 46.01 & 51.47 \\
& Multiflow
& 51.95 & 47.80 & 54.18 & 54.55 & 47.89 & 73.77 & 53.92 & 47.59 & 51.66 \\
\rowcolor{PORTAcolor}
& \textbf{PORTA}
& \textbf{55.23} & 50.05 & \textbf{55.81} & \textbf{56.50} & \textbf{51.46} & 72.13 & \textbf{56.82} & \textbf{49.76} & \textbf{54.30} \\
\bottomrule
\end{tabular}
}
\label{tab:qwenvl}
\vspace{-1.3\baselineskip}
\end{table*}
\subsection{Main Experimental Results}
\label{sec:4.2}
\textcolor{black}
{We report zero-shot performance for CLIP models pruned by PORTA and baseline methods under high sparsity levels of 0.55--0.65 across retrieval and classification benchmarks in Tab.~\ref{tab:table4a_accuracy_8metrics}. \textcolor{black}{Overall, PORTA maintains stronger performance than existing pruning baselines in high-sparsity regimes, with particularly consistent advantages over ECoFLaP, whose performance degrades markedly beyond 50\% sparsity.}
Moreover, while LLM-oriented importance formulations used by Wanda and SparseGPT may not fully align with the characteristics of vision--language representations, PORTA achieves the best or competitive results in the majority of settings, suggesting that its importance formulation is better aligned with VLM representations.}
\textcolor{black}{Quantitatively, PORTA’s relative advantage tends to become more visible as sparsity increases.} At 65\% sparsity, PORTA improves the mean performance over the eight reported metrics by 12.6\% compared to SparseGPT and by 21.5\% compared to Wanda, as shown in Tab.~\ref{tab:table4a_accuracy_8metrics}. For example, PORTA improves MSCOCO IR@5 from 35.00 to 37.32 over SparseGPT and maintains higher ImageNet accuracy at 31.38\% versus 25.96\%. On BLIP at 45\% sparsity, as reported in Tab.~\ref{tab:table3_retrieval_flickr30k}, PORTA attains a higher mean score of 90.26 than 89.95 for SparseGPT and improves TR@1 from 81.70 to 83.00, indicating reliable effectiveness beyond CLIP.\par
\textcolor{black}{We further validate PORTA beyond CLIP/BLIP in generative VLM settings. On Qwen2-VL~\cite{wang2024qwen2vl} for ScienceQA~\cite{lu2022learn} at 50\% sparsity, as shown in Tab.~\ref{tab:qwenvl}, PORTA achieves the highest average accuracy of 54.30 compared to 61.87 for the dense model. PORTA also outperforms all pruning baselines; in particular, it surpasses the second-best baseline Wanda by 0.87\%p, increasing the average accuracy from 53.43 to 54.30. PORTA achieves the best average accuracy and remains competitive across categories, with particularly strong performance in natural science with 55.23, language with 55.81, and text-based context with 56.50. These results demonstrate that our importance formulation transfers effectively to generative VLMs and visual QA tasks. Additional results on diffusion-based models and OpenFlamingo~\cite{awadalla2023openflamingo} image captioning~\cite{bernardi2016imagecaptioning} are provided in Appendix \textcolor{red}{B} and \textcolor{red}{C}, respectively.}\par







\subsection{Pruning Ratio Visualization}
\label{sec:4.3}
\begin{wrapfigure}{r}{0.55\textwidth}
  \centering
  \vspace{-22pt}
  \includegraphics[width=0.55\textwidth]{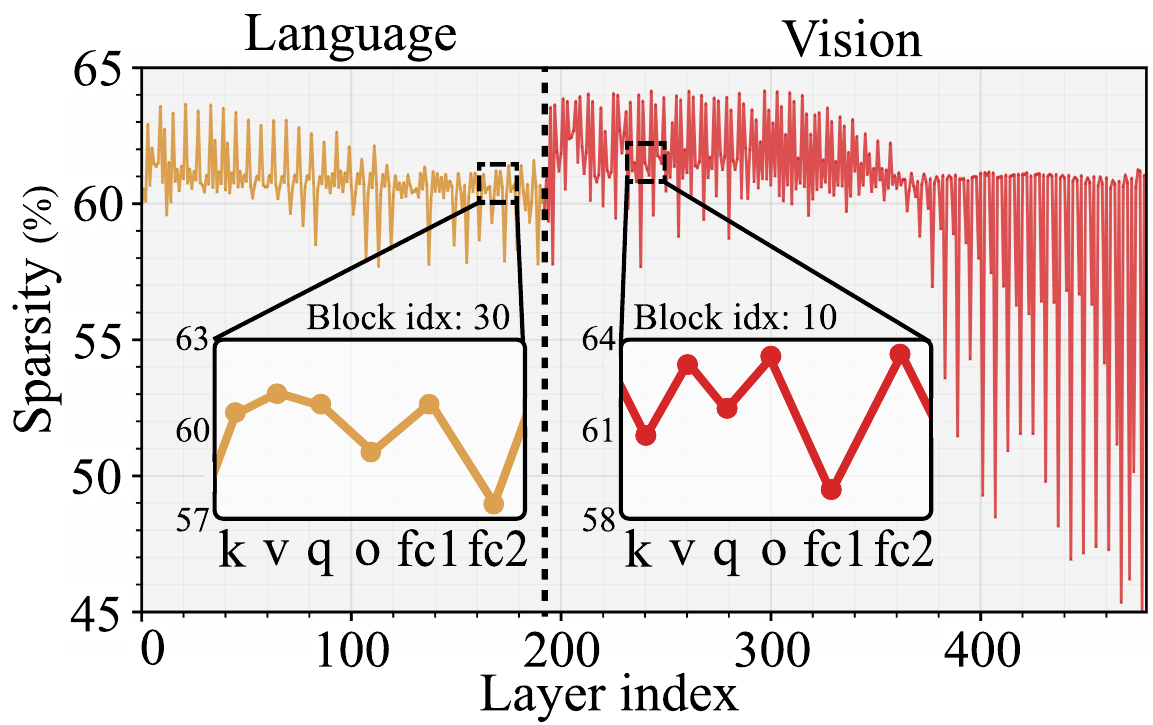}
  \caption{Layer-wise pruning ratio allocation produced by PORTA on CLIP at 60\% sparsity ($\alpha = 10$).}
  \label{fig:pruning-vlm}
  \vspace{-20pt}
\end{wrapfigure}
We visualize the layer-wise pruning ratios assigned by PORTA when pruning CLIP to a 60\% target sparsity in Fig.~\ref{fig:pruning-vlm}. \textcolor{black}{For visualization, we use a fixed $\alpha$; additional analysis is provided in Appendix \textcolor{red}{D}.} At the modality level, the language encoder exhibits relatively uniform pruning ratios in the range of 55\%--65\%, whereas the vision encoder shows substantially larger variation, spanning approximately 45\%--65\%.
Across both encoders, early layers tend to receive higher sparsity, while deeper layers are assigned lower sparsity, reflecting their greater contribution to downstream representations—an observation consistent with prior studies on representation depth in VLMs. Previous work~\cite{gandelsman2023interpreting} has shown that the final layers of CLIP models are responsible for producing the majority of its semantic representations, and~\citep{dorszewski2025colors} further demonstrates that deeper layers exhibit increasing representational complexity and encode progressively more diverse concepts.

\begin{figure}[t]
  \centering
  \includegraphics[width=\textwidth]{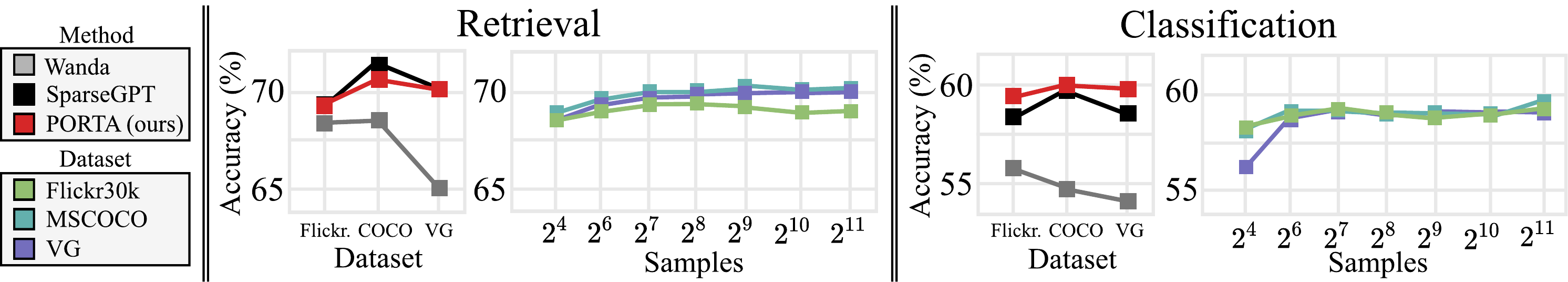}
  \caption{Effect of calibration data type, pruning method, and sample size on zero-shot performance at 60\% sparsity with CLIP.}
  \label{fig:fig4}
  \vspace{-0.3\baselineskip} 
\end{figure}
\subsection{\textcolor{black}{Calibration Data and Sample Size}}
\label{sec:4.4}
We evaluate the calibration robustness of PORTA on CLIP at 60\% sparsity under zero-shot retrieval and classification in Fig.~\ref{fig:fig4}. 
The left plot varies the calibration distribution across Flickr30K, MSCOCO, and Visual Genome~\cite{krishna2017visualgenome}. While prior methods show noticeable performance shifts as the calibration source changes, with accuracy fluctuations reaching up to about 5\%, PORTA remains substantially more stable, staying within roughly 1\% across datasets. 
This stability is consistently observed in both retrieval and classification, indicating that PORTA is robust to the choice of calibration source.

The right plot varies the number of calibration samples from $2^{4}$ to $2^{11}$. 
Except for the smallest setting, PORTA exhibits near-flat performance across a wide range of sample counts, indicating that it does not require large-scale calibration to obtain reliable importance estimates. Together, these results support a practical prune-once workflow: PORTA can be calibrated with a general-purpose image--text set and reused across heterogeneous downstream tasks without task-specific recalibration or repruning, thereby reducing deployment overhead. \textcolor{black}{Further robustness results on additional calibration datasets are provided in Appendix \textcolor{red}{E}.}
\subsection{Ablation between Importance and Sparsity Allocation}
\setlength{\intextsep}{0pt}
\setlength{\columnsep}{6pt}

\begin{wraptable}{r}{0.6\columnwidth}
\vspace{-0.9\baselineskip}

\centering
\scriptsize
\renewcommand{\arraystretch}{1.1}
\setlength{\tabcolsep}{2.7pt}

\caption{Cross-combination study of score and sparsity-allocation strategies at 60\% sparsity on CLIP. Combining the PORTA score with the PORTA ratio yields the best performance across all benchmarks.}
\label{tab:classification_60_ratio}

\resizebox{\linewidth}{!}{
\begin{tabular}{c l c c c c}
\toprule
\textbf{Sparsity} & \textbf{Score} & \textbf{Ratio} & \textbf{CIFAR10} & \textbf{CIFAR100} & \textbf{ImageNet} \\
\midrule
\multirow{8}{*}[-5pt]{\footnotesize 60\%}
& Wanda      & \multirow{2}{*}{$\times$}   & 93.20 & 70.28 & 57.35 \\
& \textbf{PORTA} &                              & 94.91 & 74.45 & 59.35 \\
\cmidrule(lr){2-6}
& Wanda      & \multirow{2}{*}{Multiflow}  & 10.03 & 1.58  & 0.11 \\
& \textbf{PORTA} &                              & 10.07 & 1.40  & 0.11 \\
\cmidrule(lr){2-6}
& Wanda      & \multirow{2}{*}{ECoFLaP}    & 93.52 & 71.16 & 61.14 \\
& \textbf{PORTA} &                              & 93.57 & 71.19 & 61.33 \\
\cmidrule(lr){2-6}
& Wanda      & \multirow{2}{*}{\textbf{PORTA}} & 94.62 & 74.29 & 59.04 \\
& \textbf{PORTA} &                              & \textbf{95.06} & \textbf{75.59} & \textbf{59.66} \\
\bottomrule
\end{tabular}
}
\vspace{-0.30\baselineskip}
\end{wraptable}

To analyze the respective contributions of the importance score and the sparsity allocation strategy, we conduct a cross-combination study under 60\% sparsity, pairing the magnitude-based importance score used by Wanda with the proposed variance-based PORTA score. 
As shown in Tab.~\ref{tab:classification_60_ratio}, replacing the magnitude-based Wanda score with the proposed variance-based PORTA score consistently improves performance across datasets, even under the same sparsity allocation scheme. Notably, when sparsity is uniformly distributed across layers, PORTA improves CIFAR-100 accuracy from 70.28 to 74.45 and ImageNet accuracy from 57.35 to 59.35, indicating that variance provides a stronger and more stable pruning signal in multimodal settings.
When comparing sparsity allocation strategies under the same importance score, 
the proposed PORTA ratio consistently improves performance over uniform allocation 
and remains competitive or superior to ECoFLaP across datasets. 

Under the PORTA score, replacing uniform allocation with the proposed PORTA ratio improves CIFAR-100 accuracy from 74.45 to 75.59 and ImageNet accuracy from 59.35 to 59.66. Similarly, under the magnitude-based score, the PORTA ratio improves CIFAR-100 from 70.28 to 74.29. Notably, combining the PORTA score with the PORTA ratio achieves the best overall performance, suggesting that variance-based importance estimation and output-variance–guided sparsity allocation complement each other. These results confirm that both components contribute to the final gains of PORTA, and that the improvement cannot be attributed to either factor alone.

\subsection{Ablation on Modeling Assumptions}\label{sec:4.6}
To examine the effect of the zero-mean assumption and the diagonal covariance approximation used in \textcolor{black}{Sec.~\ref{sec:3.3}}, we conduct an ablation study under 60\% sparsity. As shown in Tab.~\ref{tab:retrieval_ablation_60}, removing the zero-mean assumption results in negligible performance differences, by at most 0.1\%. Eliminating the diagonal approximation leads to at most a 1\% fluctuation.
\begin{table}[t]
\centering
\footnotesize
\renewcommand{\arraystretch}{1.2}
\setlength{\tabcolsep}{5pt}
\caption{Ablation of the zero-mean assumption and diagonal covariance approximation at 60\% sparsity on CLIP retrieval.}
\label{tab:retrieval_ablation_60}
\begin{tabular}{l c c c c c}
\toprule
\textbf{Method} & \textbf{TR@1} & \textbf{TR@5} & \textbf{IR@1} & \textbf{IR@5} & \textbf{Avg.} \\
\midrule
\textbf{PORTA} & 60.14 & 83.04 & 43.11 & 68.18 & 63.62 \\
PORTA w/o zero-mean & 60.20 & 83.12 & 43.11 & 68.12 & 63.64 \\
PORTA w/o diagonal approximation & 60.70 & 82.80 & 43.13 & 68.27 & 63.73 \\
\bottomrule
\end{tabular}
\vspace{-0.5\baselineskip} 
\end{table}
These results indicate that both assumptions have minimal impact on retrieval performance. 

Given that the diagonal approximation enables efficient computation and allows reuse of variance statistics from the importance estimation stage for pruning-ratio calculation, we adopt the diagonal approximation in PORTA.
\subsection{Computational Cost Analysis} \label{sec:4.7}

\begin{table}[t]
\centering
\scriptsize
\renewcommand{\arraystretch}{1.2}
\setlength{\tabcolsep}{8pt}
\caption{Pruning time comparison on CLIP. Best result is boldfaced and second-best is underlined.}
\label{tab:pruning_time}
\begin{tabular}{lccccc}
\toprule
& \textbf{PORTA} & \textbf{SparseGPT} & \textbf{Wanda} & \textbf{ECoFLaP} & \textbf{Multiflow} \\
\midrule
Pruning Time (s) & \underline{195.34} & 773.35 & \textbf{151.61} & 402.28 & 229.65 \\
\bottomrule
\end{tabular}
\vspace{-0.5\baselineskip} 
\end{table}
\textcolor{black}{One practical concern in one-shot pruning is the additional computational overhead introduced by ratio allocation across layers. To evaluate the practical efficiency of PORTA, we compare the wall-clock pruning time with representative one-shot pruning baselines under the same calibration setting.} 

\textcolor{black}{As shown in Tab.~\ref{tab:pruning_time}, PORTA completes pruning in $\sim$195s, including layer-wise ratio allocation, achieving 3.96$\times$ and 2.06$\times$ faster than SparseGPT and ECoFLaP, respectively. This efficiency is enabled by reusing the variance statistics computed for importance estimation when calculating pruning ratios via the diagonal covariance approximation discussed in Sec.~\ref{sec:4.6}.}
\section{Conclusion}
We presented \textsc{PORTA}, a prune-once, retraining-free framework for task-agnostic compression of vision--language models.
\textsc{PORTA} is motivated by the observation that commonly used magnitude-based importance measures can be strongly influenced by modality-specific activation characteristics, which may introduce modality-sensitive bias and lead to imbalanced pruning across the vision and language branches.
To obtain a more reliable indicator of representation utility, PORTA integrates activation variance statistics into importance estimation, yielding a pruning criterion that is less sensitive to modality-dependent scale and distribution effects.
\textcolor{black}{In addition, we introduce a strategy for layer-wise sparsity allocation guided by output feature variance, improving compression robustness compared with existing pruning methods under high sparsity levels.} Extensive zero-shot evaluations across representative VLM families and diverse task categories demonstrate that \textsc{PORTA} consistently outperforms strong pruning baselines at the same sparsity levels, while remaining stable under changes in calibration data and sample size.
Overall, \textsc{PORTA} advances practical VLM deployment by enabling a single pruned model to generalize across heterogeneous downstream applications without re-pruning or retraining.

\section*{Acknowledgments} 

This work was supported by the National Research Foundation of Korea (NRF) grant (RS-2025-00555943); by the Institute of Information \& Communications Technology Planning \& Evaluation (IITP)
grants (No. RS-2021-II211341; Artificial Intelligence Graduate School Program (Chung-Ang University), RS-2026-25513331; Digital Columbus Project, IITP-2026-RS-2026-25546026 and IITP-2024-RS-2024-00397085; Leading Generative AI Human Resources Development, RS-2025-02219317; AI Star Fellowship (Kookmin University), and IITP-2024-RS-2024-00417958; Global Research Support Program in the Digital Field) funded by the Korea government (MSIT).

{\small
\bibliographystyle{splncs04}
\bibliography{main}

@String(CVPR  = {IEEE Conf. Comput. Vis. Pattern Recog.})

@String(ICCV  = {Int. Conf. Comput. Vis.})

@String(ECCV  = {Eur. Conf. Comput. Vis.})

@String(NeurIPS = {Adv. Neural Inform. Process. Syst.})

@String(ICML  = {Int. Conf. Mach. Learn.})

@String(ICLR  = {Int. Conf. Learn. Represent.})

@String(CVPR  = {CVPR})

@String(ICCV  = {ICCV})

@String(ECCV  = {ECCV})

@String(NeurIPS = {NeurIPS})

@String(ICML  = {ICML})

@String(ICLR  = {ICLR})

@String(CVPR= {IEEE Conf. Comput. Vis. Pattern Recog.})

@String(ICCV= {Int. Conf. Comput. Vis.})

@String(ECCV= {Eur. Conf. Comput. Vis.})

@String(NIPS= {Adv. Neural Inform. Process. Syst.})

@String(ICLR = {Int. Conf. Learn. Represent.})

@String(NIPS  = {NeurIPS})

@inproceedings{he2020task,
  title     = {Task Agnostic Continual Learning via Meta Learning},
  author    = {He, Xu and Sygnowski, Jakub and Galashov, Alexandre and Rusu, Andrei A. and Teh, Yee Whye and Pascanu, Razvan},
  booktitle = {4th Lifelong Machine Learning Workshop at ICML},
  year      = {2020}
}

@inproceedings{jamal2019task,
  title={Task agnostic meta-learning for few-shot learning},
  author={Jamal, Muhammad Abdullah and Qi, Guo-Jun},
  booktitle={Proceedings of the IEEE/CVF conference on computer vision and pattern recognition},
  pages={11719--11727},
  year={2019}
}

@inproceedings{zhang2022bmcook,
  title={BMCook: A task-agnostic compression toolkit for big models},
  author={Zhang, Zhengyan and Gong, Baitao and Chen, Yingfa and Han, Xu and Zeng, Guoyang and Zhao, Weilin and Chen, Yanxu and Liu, Zhiyuan and Sun, Maosong},
  booktitle={Proceedings of the 2022 Conference on Empirical Methods in Natural Language Processing: System Demonstrations},
  pages={396--405},
  year={2022}
}

@inproceedings{xu2021bert,
  title={NAS-BERT: Task-agnostic and adaptive-size BERT compression with neural architecture search},
  author={Xu, Jin and Tan, Xu and Luo, Renqian and Song, Kaitao and Li, Jian and Qin, Tao and Liu, Tie-Yan},
  booktitle={Proceedings of the 27th ACM SIGKDD Conference on Knowledge Discovery \& Data Mining},
  pages={1933--1943},
  year={2021}
}

@inproceedings{wang2023task,
  title     = {Task-Agnostic Structured Pruning of Speech Representation Models},
  author    = {Wang, Haoyu and Wang, Siyuan and Zhang, Wen-Qing and Suo, Hongbin and Wan, Yulong},
  booktitle = {Proc. Interspeech 2023},
  pages     = {231--235},
  year      = {2023}
}

@inproceedings{plummer2015flickr30k,
  title={Flickr30k entities: Collecting region-to-phrase correspondences for richer image-to-sentence models},
  author={Plummer, Bryan A and Wang, Liwei and Cervantes, Chris M and Caicedo, Juan C and Hockenmaier, Julia and Lazebnik, Svetlana},
  booktitle={Proceedings of the IEEE international conference on computer vision},
  pages={2641--2649},
  year={2015}
}

@article{cao2022image,
  title={Image-text retrieval: A survey on recent research and development},
  author={Cao, Min and Li, Shiping and Li, Juntao and Nie, Liqiang and Zhang, Min},
  journal={arXiv preprint arXiv:2203.14713},
  year={2022}
}

@inproceedings{chen2025dlp,
  title     = {DLP: Dynamic Layerwise Pruning in Large Language Models},
  author    = {Chen, Yuli and Cheng, Bo and Han, Jiale and Zhang, Ying-ying and Li, Yingqi and Zhang, Shuhao},
  booktitle = {International Conference on Machine Learning (ICML)},
  year      = {2025}
}

@inproceedings{farina2024multiflow,
  title={Multiflow: Shifting towards task-agnostic vision-language pruning},
  author={Farina, Matteo and Mancini, Massimiliano and Cunegatti, Elia and Liu, Gaowen and Iacca, Giovanni and Ricci, Elisa},
  booktitle={Proceedings of the IEEE/CVF Conference on Computer Vision and Pattern Recognition},
  pages={16185--16195},
  year={2024}
}

@inproceedings{sung2024ecoflap,
  title     = {ECoFLaP: Efficient Coarse-to-Fine Layer-Wise Pruning for Vision-Language Models},
  author    = {Sung, Yi-Lin and Yoon, Jaehong and Bansal, Mohit},
  booktitle = {International Conference on Learning Representations (ICLR)},
  year      = {2024}
}

@inproceedings{lee2018,
  title={{SNIP: Single-Shot Network Pruning based on connection sensitivity}},
  author={Namhoon Lee and Thalaiyasingam Ajanthan and Philip Torr},
  booktitle=ICLR,
  year={2019}
}

@inproceedings{wang2020,
  title     = {Picking Winning Tickets Before Training by Preserving Gradient Flow},
  author    = {Wang, Chaoqi and Zhang, Guodong and Grosse, Roger},
  booktitle = {International Conference on Learning Representations (ICLR)},
  year      = {2020}
}

@inproceedings{alizadeh2022prospect,
  title     = {Prospect Pruning: Finding Trainable Weights at Initialization using Meta-Gradients},
  author    = {Alizadeh, Milad and Tailor, Shyam A. and Zintgraf, Luisa M. and van Amersfoort, Joost and Farquhar, Sebastian and Lane, Nicholas Donald and Gal, Yarin},
  booktitle = {International Conference on Learning Representations (ICLR)},
  year      = {2022}
}

@inproceedings{dejorge2021progressive,
  title     = {Progressive Skeletonization: Trimming More Fat from a Network at Initialization},
  author    = {de Jorge, Pau and Sanyal, Amartya and Behl, Harkirat Singh and Torr, Philip H. S. and Rogez, Gr{\'e}gory and Dokania, Puneet K.},
  booktitle = {International Conference on Learning Representations (ICLR)},
  year      = {2021}
}

@inproceedings{Tanaka2020PruningNN,
  title     = {Pruning Neural Networks Without Any Data by Iteratively Conserving Synaptic Flow},
  author    = {Tanaka, Hidenori and Kunin, Daniel and Yamins, Daniel L. and Ganguli, Surya},
  booktitle = {Advances in Neural Information Processing Systems (NeurIPS)},
  year      = {2020}
}

@inproceedings{wang2023ntksap,
  title     = {NTK-SAP: Improving Neural Network Pruning by Aligning Training Dynamics},
  author    = {Wang, Yite and Li, Dawei and Sun, Ruoyu},
  booktitle = {International Conference on Learning Representations (ICLR)},
  year      = {2023}
}

@inproceedings{shi2023upop,
  title={{UPop: Unified and progressive pruning for compressing vision-language transformers}},
  author={Shi, Dachuan and Tao, Chaofan and Jin, Ying and Yang, Zhendong and Yuan, Chun and Wang, Jiaqi},
  booktitle=ICML,
  year={2023}
}

@article{sun2023simple,
  title={{A Simple and Effective Pruning Approach for Large Language Models}},
  author={Sun, Mingjie and Liu, Zhuang and Bair, Anna and Kolter, J Zico},
  journal=ICLR,
  year={2024}
}

@inproceedings{
  lee2021layeradaptive,
  title={{Layer-adaptive Sparsity for the Magnitude-based Pruning}},
  author={Jaeho Lee and Sejun Park and Sangwoo Mo and Sungsoo Ahn and Jinwoo Shin},
  booktitle=ICLR,
  year={2021}
}

@inproceedings{wang2022efficientvlm,
    title = "{E}fficient{VLM}: Fast and Accurate Vision-Language Models via Knowledge Distillation and Modal-adaptive Pruning",
   author={Wang, Tiannan and Zhou, Wangchunshu and Zeng, Yan and Zhang, Xinsong},
    booktitle = "Findings of the Association for Computational Linguistics: ACL 2023",
    year = "2023",
}

@inproceedings{frantar2023sparsegpt,
  title={{SparseGPT: Massive Language Models Can Be Accurately Pruned in One-Shot}},
  author={Frantar, Elias and Alistarh, Dan},
  booktitle=ICML,
  year={2023}
}

@article{han2015deep,
  title={{Deep compression: Compressing deep neural networks with pruning, trained quantization and huffman coding}},
  author={Han, Song and Mao, Huizi and Dally, William J},
  journal=NIPS,
  year={2015}
}

@inproceedings{li2022blip,
  title={{BLIP: Bootstrapping language-image pre-training for unified vision-language understanding and generation}},
  author={Li, Junnan and Li, Dongxu and Xiong, Caiming and Hoi, Steven},
  booktitle=ICML,
  year={2022}
}

@inproceedings{li2023blip2,
  title={{BLIP-2: Bootstrapping language-image pre-training with frozen image encoders and large language models}},
  author={Li, Junnan and Li, Dongxu and Savarese, Silvio and Hoi, Steven},
  booktitle=ICML,
  year={2023}
}

@inproceedings{radford2021learning,
  title={{Learning transferable visual models from natural language supervision}},
  author={Radford, Alec and Kim, Jong Wook and Hallacy, Chris and Ramesh, Aditya and Goh, Gabriel and Agarwal, Sandhini and Sastry, Girish and Askell, Amanda and Mishkin, Pamela and Clark, Jack and others},
  booktitle=ICML,
  year={2021}
}

@inproceedings{lin2014microsoft,
  title={{Microsoft COCO: Common Objects in Context}},
  author={Lin, Tsung-Yi and Maire, Michael and Belongie, Serge and Hays, James and Perona, Pietro and Ramanan, Deva and Doll{\'a}r, Piotr and Zitnick, C Lawrence},
  booktitle=ECCV,
  year={2014},
  organization={Springer}
}

@inproceedings{papineni2002bleu,
  title={{BLEU: a Method for Automatic Evaluation of Machine Translation}},
  author={Papineni, Kishore and Roukos, Salim and Ward, Todd and Zhu, Wei-Jing},
  booktitle={Proceedings of the 40th annual meeting of the Association for Computational Linguistics},
  year={2002}
}

@inproceedings{vedantam2015cider,
  title={{CIDEr: Consensus-based Image Description Evaluation}},
  author={Vedantam, Ramakrishna and Lawrence Zitnick, C and Parikh, Devi},
  booktitle=CVPR,
  year={2015}
}

@inproceedings{nilsback2008automated,
  title={Automated flower classification over a large number of classes},
  author={Nilsback, Maria-Elena and Zisserman, Andrew},
  booktitle={Indian Conference on Computer Vision, Graphics \& Image Processing},
  year={2008},
}

@inproceedings{deng2009imagenet,
  title={Imagenet: A large-scale hierarchical image database},
  author={Deng, Jia and Dong, Wei and Socher, Richard and Li, Li-Jia and Li, Kai and Fei-Fei, Li},
  booktitle=CVPR,
  year={2009},
}

@article{williams2023does,
  title   = {How Does Calibration Data Affect the Post-Training Pruning and Quantization of Large Language Models?},
  author  = {Williams, Miles and Aletras, Nikolaos},
  journal = {arXiv preprint arXiv:2311.09755},
  year    = {2023}
}

@article{imagenet,
    author = {Russakovsky, Olga, et al},
    title = {Imagenet large scale visual recognition challenge},
    journal ={International journal of computer vision},
    year = {2015}
}

@article{sdxl,
  title   = {SDXL: Improving Latent Diffusion Models for High-Resolution Image Synthesis},
  author  = {Podell, Dustin and English, Zion and Lacey, Kyle and Blattmann, Andreas and Dockhorn, Tim and M{\"u}ller, Jonas and Penna, Joe and Rombach, Robin},
  journal = {arXiv preprint arXiv:2307.01952},
  year    = {2023}
}

@incollection{dorszewski2025colors,
  title     = {From Colors to Classes: Emergence of Concepts in Vision Transformers},
  author    = {Dorszewski, Teresa and T{\v{e}}tkov{\'a}, Lenka and Jenssen, Robert and Hansen, Lars Kai and Wickstr{\o}m, Kristoffer Knutsen},
  booktitle = {Explainable Artificial Intelligence},
  pages     = {28--47},
  year      = {2025},
  publisher = {Springer}
}

@article{gandelsman2023interpreting,
  title={Interpreting clip's image representation via text-based decomposition},
  author={Gandelsman, Yossi and Efros, Alexei A and Steinhardt, Jacob},
  journal={arXiv preprint arXiv:2310.05916},
  year={2023}
}

@inproceedings{blalock2020state,
  title     = {What is the state of neural network pruning?},
  author    = {Blalock, Davis and Ortiz, Jose Javier Gonzalez and Frankle, Jonathan and Guttag, John},
  booktitle = {Proceedings of Machine Learning and Systems (MLSys)},
  year      = {2020}
}

@article{yin2023outlier,
  title={Outlier weighed layerwise sparsity (owl): A missing secret sauce for pruning llms to high sparsity},
  author={Yin, Lu and Wu, You and Zhang, Zhenyu and Hsieh, Cheng-Yu and Wang, Yaqing and Jia, Yiling and Li, Gen and Jaiswal, Ajay and Pechenizkiy, Mykola and Liang, Yi and others},
  journal={arXiv preprint arXiv:2310.05175},
  year={2023}
}

@inproceedings{wang2024exploring,
  title={Exploring intrinsic dimension for vision-language model pruning},
  author={Wang, Hanzhang and Zhang, Jiawen and Ma, Qingyuan},
  booktitle={Forty-first International Conference on Machine Learning},
  year={2024}
}

@article{Antol2015vqa,
    author = {Antol, Stanislaw and Agrawal, Aishwarya and Lu, Jiasen and Mitchell, Margaret and Batra, Dhruv and Zitnick, C. Lawrence and Parikh, Devi},
    title = {Vqa: Visual question answering},
    journal = {Proceedings of the IEEE international conference on computer vision.},
    year = {2015}
}

@inproceedings{liu2023visual,
  title={Visual Instruction Tuning},
  author={Liu, Haotian and Li, Chunyuan and Wu, Qingyang and Lee, Yong Jae},
  booktitle={Advances in Neural Information Processing Systems},
  year={2023}
}

@article{bernardi2016imagecaptioning,
  title   = {Automatic Description Generation from Images: A Survey of Models, Datasets, and Evaluation Measures},
  author  = {Bernardi, Raffaella and Cakici, Ruket and Elliott, Desmond and Erdem, Aykut and Erdem, Erkut and Ikizler-Cinbis, Nazli and Keller, Frank and Muscat, Adrian and Plank, Barbara},
  journal = {Journal of Artificial Intelligence Research (JAIR)},
  volume  = {55},
  pages   = {409--442},
  year    = {2016}
}

@article{han2015learning,
  title={Learning both weights and connections for efficient neural network},
  author={Han, Song and Pool, Jeff and Tran, John and Dally, William},
  journal={Advances in neural information processing systems},
  volume={28},
  year={2015}
}

@article{wang2024qwen2vl,
  title={Qwen2-VL: Enhancing Vision-Language Model's Perception of the World at Any Resolution},
  author={Wang, Peng and Bai, Shuai and Tan, Sinan and Wang, Shijie and Fan, Zhihao and Bai, Jinze and Chen, Keqin and Liu, Xuejing and Wang, Jialin and Ge, Wenbin and Fan, Yang and Dang, Kai and Du, Mengfei and Ren, Xuancheng and Men, Rui and Liu, Dayiheng and Zhou, Chang and Zhou, Jingren and Lin, Junyang},
  journal={arXiv preprint arXiv:2409.12191},
  year={2024}
}

@inproceedings{lu2022learn,
  title={Learn to Explain: Multimodal Reasoning via Thought Chains for Science Question Answering},
  author={Lu, Pan and Mishra, Swaroop and Xia, Tengyu and Qiu, Liang and Chang, Kai-Wei and Zhu, Song-Chun and Tafjord, Oyvind and Clark, Peter and Kalyan, Ashwin},
  booktitle={NeurIPS},
  year={2022}
}

@inproceedings{lin2004rouge,
  title={{ROUGE}: A Package for Automatic Evaluation of Summaries},
  author={Lin, Chin-Yew},
  booktitle={Text Summarization Branches Out},
  pages={74--81},
  year={2004},
  organization={Association for Computational Linguistics}
}

@inproceedings{agrawal2019nocaps,
  title={no{C}aps: novel object captioning at scale},
  author={Agrawal, Harsh and Desai, Karan and Wang, Yufei and Chen, Xinlei and Jain, Rishabh and Johnson, Mark and Batra, Dhruv and Parikh, Devi and Lee, Stefan and Anderson, Peter},
  booktitle={Proceedings of the IEEE/CVF International Conference on Computer Vision (ICCV)},
  pages={8948--8957},
  year={2019}
}

@article{awadalla2023openflamingo,
  title={{OpenFlamingo}: An Open-Source Framework for Training Large Autoregressive Vision-Language Models},
  author={Awadalla, Anas and Gao, Irena and Gardner, Josh and Hessel, Jack and Hanafy, Yusuf and Zhu, Wanrong and Marathe, Kalyani and Bitton, Yonatan and Gadre, Samir and Sagawa, Shiori and Jitsev, Jenia and Kornblith, Simon and Koh, Pang Wei and Ilharco, Gabriel and Wortsman, Mitchell and Schmidt, Ludwig},
  journal={arXiv preprint arXiv:2308.01390},
  year={2023}
}

@techreport{krizhevsky2009cifar,
  title={Learning multiple layers of features from tiny images},
  author={Krizhevsky, Alex},
  institution={University of Toronto},
  year={2009}
}

@article{krishna2017visualgenome,
  title={Visual Genome: Connecting language and vision using crowdsourced dense image annotations},
  author={Krishna, Ranjay and Zhu, Yuke and Groth, Oliver and Johnson, Justin and Hata, Kenji and Kravitz, Joshua and Chen, Stephanie and Kalantidis, Yannis and Li, Li-Jia and Shamma, David A. and Bernstein, Michael S. and Fei-Fei, Li},
  journal={International Journal of Computer Vision},
  volume={123},
  number={1},
  pages={32--73},
  year={2017},
  publisher={Springer}
}
}

\clearpage
\appendix
\renewcommand{\thetable}{A\arabic{table}}
\renewcommand{\thefigure}{A\arabic{figure}}
\setcounter{table}{0}
\setcounter{figure}{0}
\begin{center}
{\Large\bfseries Supplementary Material for\\
Prune Once: Retraining-Free Task-Agnostic Pruning for Vision-Language Models}
\end{center}
\vspace{1.5em}
\section{Additional Comparison with Multiflow}
\label{sec:Multiflow_finetuning}
We compare PORTA with Multiflow under a unified zero-shot protocol, following the setup in Sec. \textcolor{red}{4.1}. All methods are evaluated immediately after one-shot pruning using identical calibration data and without retraining. This protocol is aligned with the main objective of PORTA: compressing a model once and deploying it directly across downstream tasks without additional optimization. 
At the same time, since Multiflow was originally proposed in a setting that includes post-pruning fine-tuning, we additionally report fine-tuning results to provide a broader and more balanced comparison. 

\par\vspace{1em} 
\begin{table}[H]
\centering
\scriptsize
\setlength{\tabcolsep}{6pt}
\renewcommand{\arraystretch}{1.2}
\setlength{\extrarowheight}{0.8pt}

\caption{Comparison with Multiflow on MSCOCO image-text retrieval at 60\% sparsity, with and without fine-tuning (FT).}
\label{tab:finetuning_results}
\begin{tabular*}{\linewidth}{@{\extracolsep{\fill}}@{}c c l c c c c@{}}
\hline
\textbf{Sparsity} & \textbf{FT} & \textbf{Method}
& \textbf{TR@1} & \textbf{TR@5} & \textbf{IR@1} & \textbf{IR@5} \\
\hline
\multirow{4}{*}{\centering 60\%}
& \multirow{2}{*}{\centering\ensuremath{\times}} & Multiflow & 0.04 & 0.16 & 0.01 & 0.08 \\
& & \textbf{PORTA(Ours)} & \textbf{60.14} & \textbf{83.04} & \textbf{43.11} & \textbf{68.18} \\
\cline{2-7}
& \multirow{2}{*}{\centering\ensuremath{\circ}} & Multiflow & 57.74 & 83.02 & 42.37 & 71.61 \\
& & \textbf{PORTA(Ours)} & \textbf{74.40} & \textbf{92.98} & \textbf{57.63} & \textbf{82.32} \\
\hline
\end{tabular*}
\end{table}

\par\vspace{1em}
\begin{table}[H]
\centering
\scriptsize
\caption{Comparison with Multiflow on MSCOCO image-text retrieval across sparsity levels without fine-tuning.}
\label{tab:Multiflow_sparsity}
\setlength{\tabcolsep}{4pt}
\renewcommand{\arraystretch}{1}
\begin{tabular}{c l c c c c}
\toprule
\multirow[c]{3}{*}[-4pt]{\textbf{Sparsity}} & \multirow[c]{3}{*}[-4pt]{\textbf{Method}} & \multicolumn{4}{c}{\textbf{Image-text retrieval}} \\
\cmidrule(lr){3-6}
& & \multicolumn{4}{c}{\textbf{MSCOCO}} \\
\cmidrule(lr){3-6}
& & \textbf{TR@1} & \textbf{TR@5} & \textbf{IR@1} & \textbf{IR@5} \\
\midrule
0\%  & Dense & 68.56 & 87.70 & 52.28 & 76.09 \\
\midrule
\multirow[c]{2}{*}{30\%} & Multiflow & 66.60 & 87.38 & 50.11 & 74.06 \\
                         & \textbf{PORTA(Ours)} & \textbf{68.32} & \textbf{88.00} & \textbf{52.28} & \textbf{75.96} \\
\midrule
\multirow[c]{2}{*}{35\%} & Multiflow & 60.72 & 83.48 & 43.33 & 67.41 \\
                         & \textbf{PORTA(Ours)} & \textbf{68.52} & \textbf{88.08} & \textbf{52.26} & \textbf{75.82} \\
\midrule
\multirow[c]{2}{*}{40\%} & Multiflow & 38.58 & 62.50 & 21.37 & 40.66 \\
                         & \textbf{PORTA(Ours)} & \textbf{68.56} & \textbf{88.24} & \textbf{51.86} & \textbf{75.67} \\
\midrule
\multirow[c]{2}{*}{45\%} & Multiflow & 5.50  & 13.78 & 1.88  & 5.24 \\
                         & \textbf{PORTA(Ours)} & \textbf{68.44} & \textbf{88.00} & \textbf{51.62} & \textbf{75.33} \\
\midrule
\multirow[c]{2}{*}{50\%} & Multiflow & 0.86  & 2.36  & 0.08  & 0.46 \\
                         & \textbf{PORTA(Ours)} & \textbf{67.90} & \textbf{87.82} & \textbf{50.69} & \textbf{74.73} \\
\bottomrule
\end{tabular}
\par\vspace{0.8em}
\end{table}

Tab.~\ref{tab:finetuning_results} shows that although Multiflow benefits substantially from fine-tuning, PORTA still yields stronger retrieval performance overall. This observation suggests that the advantage of PORTA is not limited to the retraining-free regime, but persists even when both methods are compared under a fine-tuning protocol.

To further clarify the behavior of Multiflow across sparsity levels, Tab.~\ref{tab:Multiflow_sparsity} reports additional CLIP retrieval results on MSCOCO from 30\% to 50\% sparsity without fine-tuning. At lower sparsity, both methods remain close to the dense baseline; however, as sparsity increases, Multiflow exhibits a sharp performance drop, whereas PORTA remains comparatively stable. This consistent trend helps explain why the gap in the main results becomes particularly pronounced at high sparsity, and supports that the observed behavior reflects differing sparsity robustness rather than an implementation artifact.

\section{Additional Evaluation on Diffusion}
\label{sec:additional_diffusion}
\begin{figure}[H]
    \centering
    \includegraphics[width=0.85\linewidth]{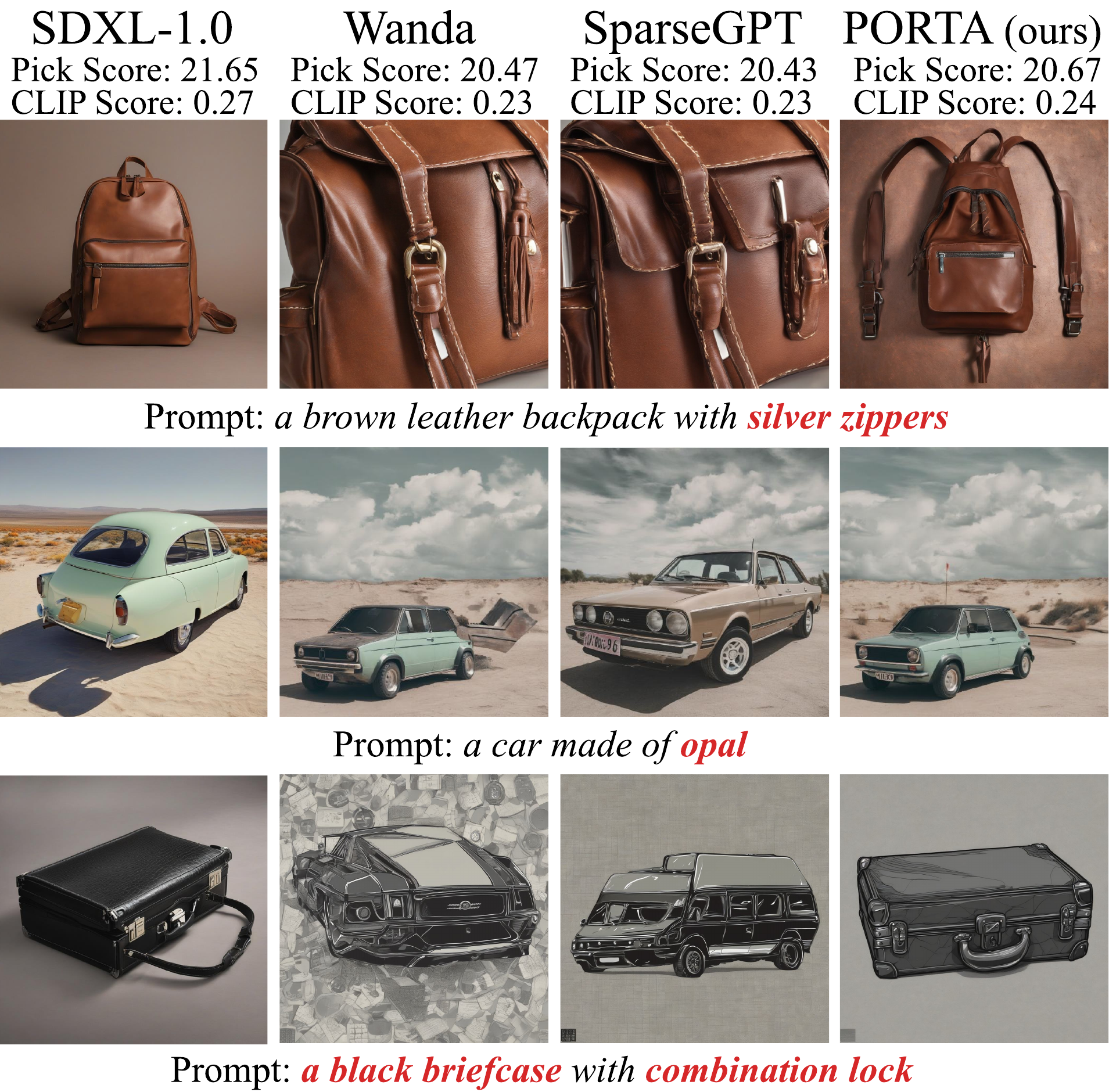}
    \caption{Qualitative comparison of zero-shot text-to-image generation results on SDXL-1.0 under different pruning methods.}
    \label{fig:fig5}
    \vspace{1em}
\end{figure}

As discussed in Sec. \textcolor{red}{4.2}, we further evaluate PORTA on text-to-image diffusion with SDXL 1.0~\cite{sdxl}, where the text encoder plays a critical role in preserving prompt semantics. We prune the OpenAI CLIP-ViT-bigG text encoder to 65\% sparsity and the CLIP ViT-L/14 text encoder to 30\% sparsity. Since diffusion generation relies only on the text encoder, we disable PORTA’s ratio mechanism—designed to allocate sparsity across modalities—to enable a fair comparison with Wanda and SparseGPT, both of which use uniform layer sparsity.

Across all prompts shown in Fig.~\ref{fig:fig5}, PORTA achieves a higher PickScore than the baselines. For example, PORTA obtains a PickScore of 20.67, whereas Wanda and SparseGPT achieve 20.47 and 20.43, respectively, while maintaining competitive CLIPScore. Qualitatively, PORTA preserves fine-grained semantic attributes—such as ``silver zippers,'' ``opal,'' and ``combination lock''—more accurately than the comparison methods, whose outputs often omit or distort these details.

Together, these results show that PORTA remains effective not only on CLIP-based downstream tasks, but also when applied to diffusion text encoders. This supports the broad applicability of PORTA across diverse multimodal models.




\section{Additional Evaluation on OpenFlamingo}
\label{sec:additional_openflamingo}
\begin{table}[h]
\centering
\footnotesize
\renewcommand{\arraystretch}{1.1}
\setlength{\tabcolsep}{5pt}
\caption{OpenFlamingo results on MSCOCO image captioning at 30\% sparsity.}
\label{tab:flamingo}
\begin{tabular}{c c c c c c}
\toprule
\multirow{3}{*}[-5pt]{\textbf{Sparsity}} & \multirow{3}{*}[-5pt]{\textbf{Method}} & \multicolumn{4}{c}{\textbf{Image Captioning}} \\
\cmidrule(lr){3-6}
& & \multicolumn{4}{c}{\textbf{MSCOCO}} \\
\cmidrule(lr){3-6}
& & \textbf{BLEU@1} & \textbf{BLEU@2} & \textbf{ROUGE-L} & \textbf{CIDEr} \\
\midrule
0\% & \textbf{Dense} & 44.52 & 27.29 & 28.79 & 37.87 \\
\midrule
\multirow{5}{*}{30\%}
& Wanda     & 32.01 & 18.04 & 22.91 & 22.11 \\
& SparseGPT & 41.71 & 25.86 & 27.54 & 34.75 \\
& Multiflow & 6.62  & 1.42  & 15.14 & 1.45  \\
& ECoFLaP   & 41.10 & 24.46 & 27.48 & 31.77 \\
& \textbf{PORTA(Ours)} & \textbf{43.58} & \textbf{26.61} & \textbf{28.84} & \textbf{37.02} \\
\bottomrule
\end{tabular}
\par\vspace{0.8em}
\end{table} As discussed in Sec. \textcolor{red}{4.2}, we further evaluate PORTA on image captioning~\cite{bernardi2016imagecaptioning} with OpenFlamingo~\cite{awadalla2023openflamingo}, a generative VLM whose architecture differs from the Qwen2-VL~\cite{wang2024qwen2vl} model used in the main text. We prune OpenFlamingo on MSCOCO~\cite{lin2014microsoft} image captioning at 30\% sparsity and report BLEU@1/2~\cite{papineni2002bleu}, ROUGE-L~\cite{lin2004rouge}, and CIDEr~\cite{vedantam2015cider}. \par
~\textcolor{black}{As shown in Tab.~\ref{tab:flamingo}, PORTA nearly preserves the dense performance of 37.87, achieving a CIDEr score of 37.02, and outperforms all pruning baselines. In particular, PORTA improves CIDEr over Wanda from 22.11 to 37.02, a 67.4\% relative gain.
Together, these results show that PORTA is effective not only in the VQA setting of the main text but also on image captioning with a different generative VLM family, supporting the broad applicability of PORTA across diverse multimodal models.}

\section{Pruning Ratio Strength}
\label{sec:Pruning Ratio Strength}

\setlength{\columnsep}{8pt}
\begin{wraptable}[8]{r}{0.5\columnwidth}
\vspace{-1\baselineskip} 
\centering
\scriptsize
\renewcommand{\arraystretch}{0.93}
\setlength{\tabcolsep}{8pt}
\caption{Zero-shot performance across different $\alpha$ under varying sparsity. Best within each sparsity is boldfaced.}
\label{tab:table8}
\begin{tabular}{ccccc}
\toprule
\multirow{2}{*}[-1pt]{$\boldsymbol{\alpha}$} & \multicolumn{4}{c}{\textbf{Sparsity}} \\
\cmidrule(lr){2-5}
& \textbf{50\%} & \textbf{55\%} & \textbf{60\%} & \textbf{65\%} \\
\midrule
1  & 96.34 & 94.76 & 94.70 & 89.03 \\
5  & \textbf{96.44} & 94.89 & 94.89 & 89.09 \\
10 & 96.40 & \textbf{96.40} & \textbf{95.06} & 89.36 \\
30 & 96.34 & 95.58 & 94.76 & \textbf{90.49} \\
\bottomrule
\end{tabular}
\vspace{-0.5\baselineskip}
\end{wraptable}
As discussed in Sec. \textcolor{red}{4.3}, the pruning ratio in PORTA is controlled by a hyperparameter that adjusts the allowable range of ratio variation across layers. This mechanism reflects the observation that higher target sparsity requires larger variation in layer-wise allocation range, whereas lower sparsity favors a narrower allocation range—an effect also reported in recent work~\cite{chen2025dlp}.
Tab.~\ref{tab:table8} presents CIFAR-10 classification results using CLIP, showing that larger $\alpha$ tends to be preferred as sparsity becomes more stringent.
\section{Calibration Robustness}
\label{sec:calibration_robustness}
\setlength{\columnsep}{5pt}
\begin{wraptable}[6]{r}{0.48\linewidth}
\vspace{-10pt} 
\centering
\scriptsize
\caption{Calibration robustness at 60\% sparsity.}
\setlength{\tabcolsep}{10pt}
\renewcommand{\arraystretch}{1.0}
\begin{tabular}{@{}lcc@{}}
\toprule
\textbf{Calib.} & \textbf{ImageNet} & \textbf{Retrieval} \\
\midrule
COCO & 59.30$\pm$.06 & 69.76$\pm$.53 \\
NoCaps & 60.22$\pm$.45 & 69.50$\pm$.30 \\
PASCAL-50S & 59.57$\pm$.29 & 69.25$\pm$.36 \\
\bottomrule
\end{tabular}
\label{tab:calib_data_robustness}
\end{wraptable}
As discussed in Sec. \textcolor{red}{4.4}, to verify that PORTA does not depend on the specific distribution of the calibration source, we further evaluate it at 60\% sparsity using NoCaps~\cite{agrawal2019nocaps} and PASCAL-50S~\cite{vedantam2015cider}, which differ from the Flickr-style calibration data used in the main experiments.
As shown in Tab.~\ref{tab:calib_data_robustness}, PORTA remains stable across sample sizes from 64 to 512, with standard deviations below 0.45 for ImageNet top-1 accuracy and 0.36 for retrieval, comparable to COCO. These results indicate that PORTA is robust across diverse calibration distributions without relying on a specific calibration source.

\section{Structured pruning}
\label{sec:structured_pruning}

\setlength{\columnsep}{3pt}
\begin{wraptable}{r}{0.4\linewidth}
\vspace{-12pt} 
\centering
\scriptsize
\renewcommand{\arraystretch}{0.95}
\setlength{\tabcolsep}{3pt}
\caption{Latency and memory.}
\label{tab:latency}
\begin{tabular}{lcc}
\toprule
 & \textbf{Dense} & \textbf{2:4} \\
\midrule
Latency (ms) & 17.2 & \textbf{10.3} ($\times$1.67) \\
Memory (GB)  & 4.71 & \textbf{2.65} ($\times$1.78) \\
\bottomrule
\end{tabular}
\vspace{0pt} 
\end{wraptable}
\textcolor{black}{We extend PORTA to 2:4 semi-structured pruning on CLIP-ViT-bigG for practical latency and memory efficiency. 
As shown in Tab.~\ref{tab:latency}, at batch size 1 the encoder latency improves from 17.2 ms to 10.3 ms, a 1.67$\times$ speedup, and the memory footprint is reduced from 4.71 GB to 2.65 GB, a 1.78$\times$ reduction.}
\begin{table}[h]
\vspace{4pt}
\centering
\scriptsize
\renewcommand{\arraystretch}{1.0}
\setlength{\tabcolsep}{6pt}
\caption{Zero-shot accuracy under 2:4 sparsity on CLIP-ViT-bigG.}
\label{tab:2to4_accuracy}
\resizebox{\textwidth}{!}{
\begin{tabular}{lcccccc}
\toprule
\multirow{2}{*}[-1pt]{\textbf{Method}} & \multirow{2}{*}[-1pt]{\textbf{Sparsity}} & \multicolumn{5}{c}{\textbf{Accuracy}} \\
\cmidrule(lr){3-7}
& & \textbf{CIFAR-100} & \textbf{ImageNet} & \textbf{Flowers102} & \textbf{Retrieval} & \textbf{Average} \\
\midrule
Dense      & 0\%  & 87.50 & 78.45 & 80.19 & 76.90 & 80.76 \\
\midrule
Wanda      & \multirow{3}{*}{2:4} & 72.35 & 52.38 & 43.19 & 66.19 & 58.53 \\
SparseGPT  & & 72.63 & 56.71 & 53.14 & \textbf{69.55} & 63.01 \\
\textbf{PORTA(Ours)}& & \textbf{73.80} & \textbf{57.33} & \textbf{56.27} & \underline{69.54} & \textbf{64.24} \\
\bottomrule
\end{tabular}
}
\vspace{4pt}
\end{table}

\textcolor{black}{Beyond these efficiency gains, as shown in Tab.~\ref{tab:2to4_accuracy}, PORTA achieves an average accuracy of 64.24 under 2:4 sparsity, surpassing SparseGPT at 63.01 and Wanda at 58.53, the highest average accuracy among the pruning methods. In particular, PORTA outperforms all baselines on CIFAR-100 and Flowers102, reaching 73.80 and 56.27 respectively, and also attains the highest ImageNet accuracy at 57.33. This demonstrates that PORTA's variance-based importance estimation remains effective under the 2:4 structured constraint, and that the structured sparsity pattern does not compromise its performance.}

\section{Modality-agnostic Variance-based Metrics}
\label{sec:modality_agnostic}

\begin{table}[H]
\centering
\scriptsize
\caption{Classification results at 65\% sparsity under different pruned branches.}
\label{tab:classificationC}
\setlength{\tabcolsep}{2pt}
\renewcommand{\arraystretch}{1.2}
\setlength{\extrarowheight}{0.8pt}

\begin{tabular*}{\linewidth}{@{\extracolsep{\fill}}@{}c
  >{\centering\arraybackslash}m{1.25cm}
  l c c c c c@{}}
\hline
\textbf{Sparsity} & \textbf{Pruned} & \textbf{Method}
& \textbf{CIFAR10} & \textbf{CIFAR100} & \textbf{ImageNet} & \textbf{Flowers102} & \textbf{Avg} \\
\hline
\multirow{6}{*}{\centering 65\%}
& \multirow{2}{*}{\centering Both}
& Wanda          & 69.81 & 28.86 & 30.06 & 17.30 & 36.50 \\
&
& \textbf{PORTA w/o ratio} & \textbf{90.49} & \textbf{62.23} & \textbf{31.38} & \textbf{18.27} & \textbf{50.59} \\
\cline{2-8}
& \multirow{2}{*}{\centering Text}
& Wanda          & 97.72 & 73.89 & 35.21 & 23.50 & 57.58 \\
&
& \textbf{PORTA w/o ratio} & \textbf{97.72} & \textbf{77.89} & \textbf{44.24} & \textbf{28.98} & \textbf{62.20} \\
\cline{2-8}
& \multirow{2}{*}{\centering Vision}
& Wanda          & 73.66 & 31.67 & 63.02 & 56.92 & 56.31 \\
&
& \textbf{PORTA w/o ratio} & \textbf{87.64} & \textbf{61.65} & \textbf{66.27} & \textbf{61.14} & \textbf{69.17} \\
\hline
\end{tabular*}
\end{table}


To further support the discussion in Sec. \textcolor{red}{3.1}, we quantitatively compare pruning results obtained using magnitude-based and variance-based importance scores in multimodal pruning. Specifically, we apply Wanda and PORTA to CLIP under a uniform pruning ratio and report module-wise classification performance when pruning the vision encoder, the text encoder, or both modules in Tab.~\ref{tab:classificationC}. PORTA, which uses a variance-based importance score, consistently outperforms Wanda, which relies on magnitude-based importance, across all pruning settings. When both modules are pruned, PORTA improves the average accuracy from 36.50 to 50.59, corresponding to a relative improvement of 38.60\%. In particular, when only the text encoder is pruned, the average accuracy increases from 57.58 to 62.20, resulting in a modest improvement of 8.02\%. In contrast, when only the vision encoder is pruned, the average accuracy increases from 56.31 to 69.17, yielding a substantially larger improvement of 22.84\%, which is notably greater than the gain observed in the text-only pruning setting. This result is consistent with the analysis in Sec. \textcolor{red}{3.1} and demonstrates that, in multimodal settings, a variance-based importance score provides a more effective pruning signal than magnitude-based criteria.

\end{document}